\documentclass[letterpaper]{article} 
\usepackage[preprint]{aaai2027}  
\usepackage[hyphens]{url}  
\usepackage{graphicx} 
\usepackage{pifont}
\def\UrlFont{\rm}  
\usepackage{natbib}  
\usepackage{caption} 
\usepackage{algorithm}
\usepackage{algorithmic}
\usepackage[ruled,vlined,linesnumbered,algo2e]{algorithm2e}
\usepackage{framed}
\usepackage{float}
\usepackage{amsmath}
\usepackage{amssymb}

\usepackage{booktabs}
\usepackage{tabularx}
\newcolumntype{Y}{>{\centering\arraybackslash}X}

\usepackage{array}
\usepackage{multirow}
\usepackage{bibunits}

\definecolor{DuLBEGray}{RGB}{242,243,245}
\definecolor{DuLBEBlue}{RGB}{229,239,250}
\definecolor{DuLBEYellow}{RGB}{252,246,220}
\definecolor{DuLBEOrange}{RGB}{180,70,0}

\makeatletter
\newcommand{\DuLBEModuleTitle}[1]{%
    \algocf@seteveryparnl{\relax}%
    \textbf{#1}\par
    \algocf@linesnumbered}
\newcommand{\DuLBEAppendixTitle}{%
    \twocolumn[
    \vbox{%
        \hsize\textwidth
        \linewidth\hsize
        \vskip 0.625in minus 0.125in
        \centering
        {\LARGE\bf Dual-Mode Low-Rank Learner with
        Bridge-Prototype Ensemble for Vision-Language
        Class-Incremental Learning\par}
        \vskip 0.1in
        {\Large\bf Appendix\par}
        \vskip 0.1in
        {\Large\bf
        Chiyuan He\textsuperscript{\rm 1},
        Zihuan Qiu\textsuperscript{\rm 1},
        Fanman Meng\textsuperscript{\rm 1,\ding{41}},
        Chao Wang\textsuperscript{\rm 2},
        Liangjiang Chen\textsuperscript{\rm 1},
        Linfeng Xu\textsuperscript{\rm 1},
        Qingbo Wu\textsuperscript{\rm 1},
        Hongliang Li\textsuperscript{\rm 1}\par}
        \vskip .2em
        {\normalsize
        \textsuperscript{\rm 1}University of Electronic Science and Technology of China, Chengdu, China\par
        \textsuperscript{\rm 2}Qiyuan Lab, Beijing, China\par
        \{cyhe,zihuanqiu,202522011613\}@std.uestc.edu.cn\par
        \{lfxu,qbwu,hlli\}@uestc.edu.cn,\quad
        w-c15@tsinghua.org.cn\par}
        \vskip 1em
    }]
    \insert\footins{\noindent\footnotesize
        \ding{41}\ Corresponding author: fmmeng@uestc.edu.cn.\quad
        Preprint. Work in progress.\par}}
\makeatother

\title{Dual-Mode Low-Rank Learner with Bridge-Prototype Ensemble for Vision-Language Class-Incremental Learning}

\author{
    Chiyuan He\textsuperscript{\rm 1},
    Zihuan Qiu\textsuperscript{\rm 1},
    Fanman Meng\textsuperscript{\rm 1,\ding{41}},
    Chao Wang\textsuperscript{\rm 2},
    Liangjiang Chen\textsuperscript{\rm 1},
    Linfeng Xu\textsuperscript{\rm 1},
    Qingbo Wu\textsuperscript{\rm 1},
    Hongliang Li\textsuperscript{\rm 1}
}
\affiliations{
    \textsuperscript{\rm 1}University of Electronic Science and Technology of China, Chengdu, China\\
    \textsuperscript{\rm 2}Qiyuan Lab, Beijing, China\\
    \{cyhe,zihuanqiu,202522011613\}@std.uestc.edu.cn\\
    \{lfxu,qbwu,hlli\}@uestc.edu.cn,\quad
    w-c15@tsinghua.org.cn
}

\copyrighttext{\ding{41}\ Corresponding author: fmmeng@uestc.edu.cn.\quad
Preprint. Work in progress.}

\begin{document}

\maketitle

\begin{abstract}
Benefiting from transferable visual-textual alignment, CLIP has been widely adopted for class-incremental learning (CIL). However, existing learners either
repeatedly update components shared across tasks, leading to knowledge overwriting, or overly isolate new-task updates, hindering the reuse of CLIP's
transferable knowledge and limiting plasticity. Moreover, the text-based or bimodal classifier designs still fail to effectively integrate complementary information from the visual and textual
modalities. To address these challenges, we introduce DuLBE, which couples dual-mode low-rank learning with a bridge-prototype ensemble classifier for exemplar-free CIL. DuLBE allocates two visual low-rank update modes according to the gradient demand and uses gradient routing to coordinate them: a compact and rewritable shared mode is selected from historically occupied visual directions to reuse transferable knowledge, while residual modes provide low-interference channels for task-specific variations. Building on the resulting stable inter-modal structure, we further construct geodesic bridges between visual prototypes and text embeddings on the unit hypersphere, and ensemble reliable bridge prototypes to compensate for the modality-gap limitations of textual decision boundaries. Extensive experiments under multiple settings show that DuLBE achieves state-of-the-art CIL performance while retaining the high parameter efficiency of low-rank tuning.
\end{abstract}

\begin{figure}[t]
    \centering
    \includegraphics[width=\linewidth]{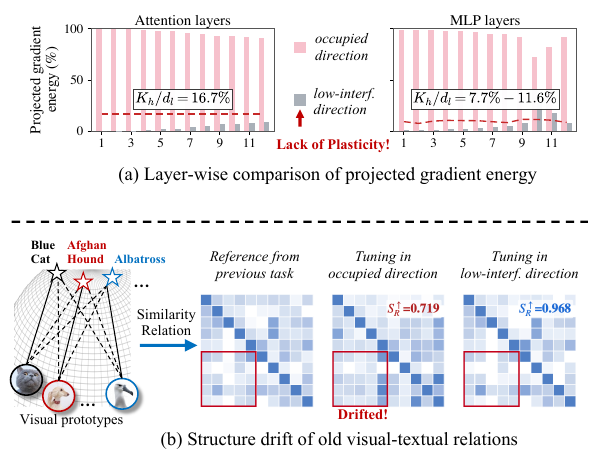}
\caption{Motivating observations.
(a) New-task cross-modal gradients retain substantial projection energy in
historically occupied directions ($K_h=128$) across the visual tower.
(b) Tuning occupied directions causes greater old-class relation-structure
drift than tuning low-interference directions. $S_R$ denotes rank-structure similarity to the old-task
relation matrix, whose entries are cosine similarities between old-class
visual prototypes and corresponding text embeddings.}
\label{fig:motivation}
\end{figure}

\section{Introduction}
Benefiting from large-scale pre-training over aligned
vision-text data, vision-language models (VLMs) like
CLIP~\cite{radford2021learning} can use natural-language text to organize
the continually expanding class space of downstream class-incremental
learning (CIL). The central challenge is to balance stability and
plasticity: when previous data are unavailable or strictly limited, the
model must preserve knowledge from old classes while
continually incorporating knowledge of new classes.

Parameter-efficient tuning methods for pre-trained CLIP commonly freeze the
backbone and introduce a small number of trainable parameters, such as prompts,
adapters, or low-rank factors~\cite{zhou2022learning,gao2024clip,hu2022lora}.
These methods retain the general alignment knowledge acquired during
pre-training while adapting the model to downstream tasks. Building on this
adaptation paradigm, recent continual learning methods introduce global
constraints~\cite{zheng2023preventing,yu2024exploiting,wu2025synthetic,he2026segpcl}, or tailored update strategies ~\cite{zhang2024overcoming,huang2025mind} to continually adapt CLIP. Nevertheless, these methods still repeatedly
optimize trainable components shared across tasks. Without old-task data,
new-task supervision can progressively overwrite the previously acquired
knowledge encoded in these shared components.

Another line of work isolates task knowledge to reduce interference from
new-task learning, with particular attention to the more fragile visual
tower~\cite{zhou2025engine,li2025bofa,kang2025dmnsp,peng2025gnsp,qiu2026null}.
Within this line, strategies based on orthogonal update or null-space
learning reduce interference by steering new-task updates away from the principal
directions occupied by old-task representations. However, our study reveals that avoiding these
directions may be overly restrictive in CLIP. As shown in Figure~\ref{fig:motivation}(a), a substantial portion of the
new-task cross-modal gradient energy in CIL is projected onto these
occupied directions (i.e., the top-$K_h$ principal directions of old-task
representations), well above the dimensional baseline $K_h/d_l$ expected
for a random $K_h$-dimensional subspace, where $d_l$ is the input
dimension of layer $l$.
We attribute this concentration to the way CLIP organizes
visual and textual concepts in a shared general-purpose alignment space,
where different classes reuse common visual patterns (e.g., wheel-related
features shared by concepts: trucks and bicycles). Excluding these
directions would therefore discard transferable write-in signals and
restrict alignment plasticity. However, such transferability does not
make them freely rewritable. Figure~\ref{fig:motivation}(b) shows that
tuning in the occupied directions induces substantially greater drift in
old-class visual-textual relations than tuning in low-interference directions
(i.e., directions selected from the orthogonal complement
of their span). Without old data, aligning towards new textual objective can shift these shared visual patterns, thereby disrupting the old semantic relation structure and ultimately leading to forgetting. Taken together, these observations reveal that occupied directions support transfer but require controlled rewriting, whereas
low-interference directions enable safer adaptation but offer limited
plasticity.

To address this bottleneck, we propose \textbf{DuLBE}
(\textbf{Du}al-mode \textbf{L}ow-rank learner with
\textbf{B}ridge-prototype \textbf{E}nsemble), a collaborative CLIP-based
CIL framework. DuLBE performs current-task-driven, layer-wise allocation
of the low-rank subspace for adaptation in CLIP's visual tower. At each adapted
layer, it decomposes the prospective gradient according to
historical visual-mode occupation. Following the principle of maximizing
gradient projection energy, DuLBE selects one compact rewritable shared
mode from occupied directions and allocates a small set of
low-interference residual modes to the remaining task-specific demand.
Gradient routing further assigns distinct responsibilities to the two
modes: the shared mode reuses a high-demand historical direction under
semantic-structure regularization, whereas the residual modes capture
task-specific variations with negligible interference to previous tasks.
This dual-mode learner transfers previously established
cross-modal structures while injecting new knowledge with a small
rank overhead. Building on the resulting stable inter-modal structure, we
further introduce a bridge-prototype ensemble classifier. It constructs bridges between visual prototypes and text embeddings on
the unit hypersphere through geodesic interpolation and uses bridge reliability to refine the decision
boundary, flexibly combining textual semantics with adapted visual
evidence.

In summary, our contributions include: (1) we propose DuLBE, a
dual-mode collaborative low-rank continual learner that allocates
rewritable and low-interference sub-modes in CLIP's visual tower, together
with a gradient-routing mechanism for stable yet plastic adaptation. (2)
we introduce a bridge-prototype ensemble classifier that leverages
stable inter-modal geodesic structures to compensate for the limitations
of a single text-based decision boundary. (3) extensive experiments
demonstrate that DuLBE achieves state-of-the-art
performance with modest trainable overhead.

\section{Related Work}

\subsection{Class-Incremental Learning}
Class-incremental learning (CIL) requires a model to learn new classes
sequentially while recognizing all seen classes without task identities.
Traditional CIL methods mitigate forgetting by storing exemplars,
distilling old responses, or allocating task-specific parameters~\cite{
delange2021continual,rebuffi2017icarl,li2017learning,douillard2020podnet}.
These strategies establish the stability-plasticity trade-off, but they
often rely on stored data, explicit old-task supervision, or expanding
capacity.

CLIP-based CIL shifts the focus from learning visual representations from
scratch to continually optimizing visual-textual
alignment.
Existing methods mainly exploit this prior from three perspectives. First,
external knowledge, textual priors, or language-guided concept
descriptions are introduced to enrich class semantics and optimize the
continual learning objective~\cite{
zhou2025engine,he2025harnessing,yin2024adapter,huang2024rapf,
yu2025language}. Second, several studies protect the global cross-modal
semantic structure by preserving modality-gap properties, relative
semantic relations, or geometric consistency during continual
adaptation~\cite{yu2024exploiting,he2025harnessing,hu2025hierarchical,liang2025boosting,he2026segpcl,gong2026learning}. Third,
interference-isolation methods either expand and freeze task-specific
projection heads~\cite{zhou2025proof,zhou2025engine,liang2025boosting}, or constrain new-task updates
in low-interference directions~\cite{li2025bofa,kang2025dmnsp,qiu2025mingle}. These studies
demonstrate the value of CLIP's semantic structure, but they either keep
updating shared trainable modules or mainly avoid risky
directions.

\subsection{LoRA-Based Continual Learning}
Low-Rank Adaptation (LoRA) ~\cite{hu2022lora} freezes pretrained weights and parameterizes updates with low-rank
factors, which makes it suitable for compact continual
adaptation. However, a small-rank update can still interfere with old
knowledge if its directions overlap with old-task-sensitive regions.
Recent LoRA-based continual learning methods therefore control the update by orthogonalization~\cite{wang2023orthogonal}, interference-free subspace allocation~\cite{liang2024inflora,luo2026keeplora}, energy-driven subspace decomposition~\cite{he2026loda} or predefined functional LoRA insertion~\cite{he2025cllora}. These methods establish
low-rank subspace control as an effective way to balance stability and
plasticity. In contrast, DuLBE uses cross-modal gradient demand and mode occupation to decide which shared directions are rewritable and which residual directions should be included.

\begin{figure*}[t]
    \centering
    \includegraphics[width=0.95\textwidth]{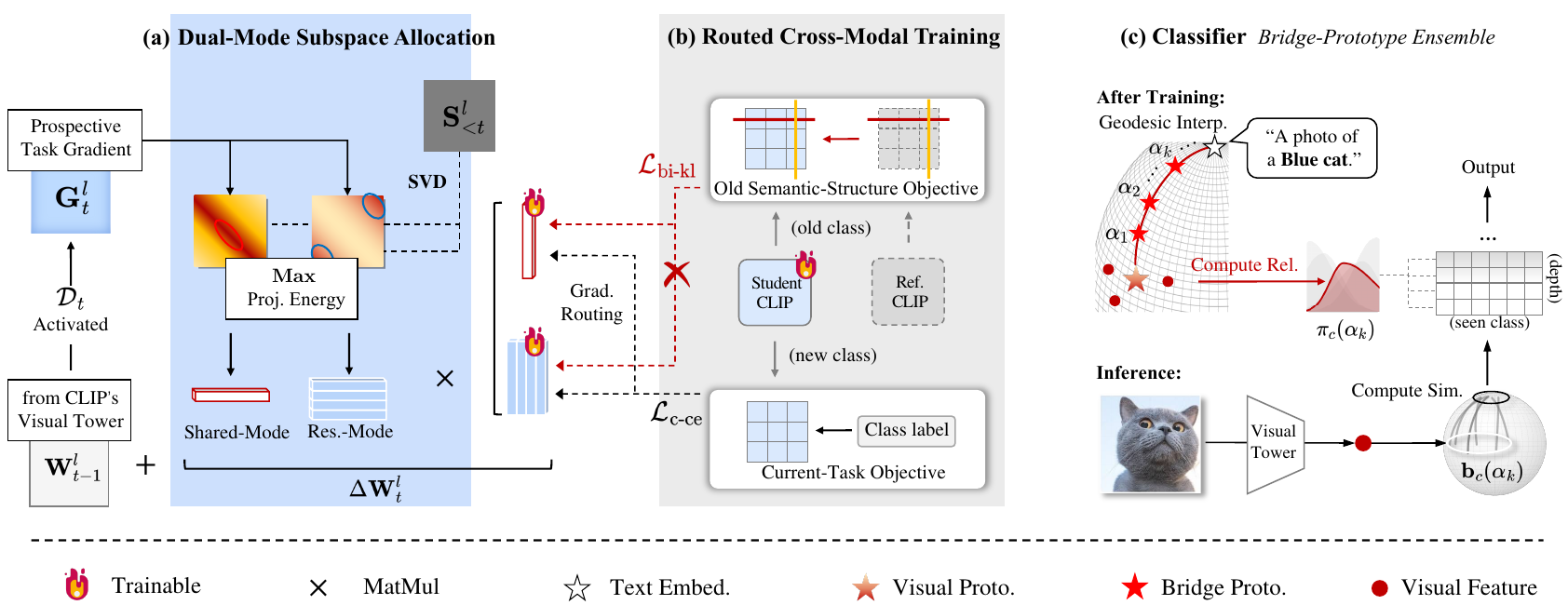}
\caption{Overview of \textbf{DuLBE}.
(a) Dual-mode allocation uses the prospective task gradient to form a
compact shared mode within historically occupied directions and a
low-interference residual mode outside them.
(b) Routed cross-modal training directs gradients from the current-task
objective to both learners, while routing gradients from the
semantic-structure preservation objective only to the shared learner.
(c) Classifier: it builds visual-text geodesic bridges, estimates
bridge-depth reliability after each task, and ensembles reliable bridge
prototypes for inference.}
    \label{fig:pipeline}
\end{figure*}

\section{Methodology}

\subsection{Preliminaries}

We consider exemplar-free class-incremental learning (CIL) with a pre-trained
CLIP model. The learner receives a sequence of tasks
$\{\mathcal D_t\}_{t=1}^{N}$, where task $t$ introduces a disjoint class
set $\mathcal C_t$ satisfying $\mathcal C_t\cap\mathcal C_k=\emptyset$
for $t\neq k$. After finishing task $t$, the model is evaluated over all
seen classes $\mathcal C_{\leq t}=\bigcup_{k=1}^{t}\mathcal C_k$ without
task identities, and no raw data from previous tasks are retained.

CLIP performs classification by comparing an image representation with
text embeddings generated from class-name prompts, e.g.,
A photo of a \{CLASS\}. Given the normalized visual feature $\mathbf z_i$ of image $x_i$ and
class-text embedding $\mathbf e_c$, the CLIP logit is
$\ell_i(c)=\tau\mathbf z_i^\top\mathbf e_c$, where $\tau$ is the logit scale. This language-defined
classifier naturally supports an expanding class space, but it also
makes continual visual adaptation delicate: the visual tower must
learn new classes while remaining compatible with both the textual
decision space and the cross-modal relations formed by previous
tasks.

\subsection{Overview}

As shown in Figure~\ref{fig:pipeline}, DuLBE integrates three components. First, it
allocates the visual low-rank update space using the current-task prospective gradient. Within the historically
occupied subspace, the direction with the strongest gradient demand forms
a compact rewritable shared mode, while dominant directions outside this
subspace form low-interference residual modes for task-specific knowledge.
Second, the shared and residual mode learners are trained collaboratively
through mode-specific gradient routing, which assigns different knowledge
roles: the current-task CE updates both learners, while
semantic-structure preservation regularizes only the shared learner,
encouraging controlled knowledge reuse and residual task-specific
plasticity. Third, DuLBE forms a reliability-guided bridge-prototype
ensemble by estimating visual-text geodesic bridges after each
task and aggregating reliable bridge prototypes for classification.

\subsection{Gradient-Demand Dual-Mode Subspace Allocation}

 A single current-task
update may overwrite old visual-text relations. Some interference-aware strategies restrict new updates to orthogonal or
null-space directions~\cite{wang2023orthogonal,liang2024inflora}, reducing
forgetting but also blocking directions that could remain reusable in
CLIP's shared semantic space. Dual-branch designs introduce shared and
task-specific components~\cite{he2025cllora,he2026loda}, but their allocation is
often predefined rather than determined by the current write-in demand. We instead allocate the visual low-rank subspace from the perspective of the
current cross-modal gradient. For each adapted visual layer $l$, a
prospective gradient is computed on current-task data $\mathcal D_t$:
\begin{equation}
    \mathbf G_t^l
    =
    \nabla_{\mathbf W_{t-1}^l}
    \mathcal L_{\mathrm{c\text{-}ce}}(\theta_{t-1};\mathcal D_t).
\end{equation}
This gradient is used only to plan the low-rank update basis, not as a
direct parameter update. It indicates which input directions are activated
by the current cross-modal CE objective in the layer $l$.

\paragraph{Historical Visual-Mode Occupation.}
The historical visual support is estimated from the second-order
statistics of layer inputs. For a previous task $t'$, the cached input tokens of the layer $l$ are row-stacked as
$\mathbf X_{t'}^l\in\mathbb R^{n_{t'}^l\times d_l}$, where $n_{t'}^l$ is
the number of cached tokens and $d_l$ is the input dimension. The task-wise and
accumulated occupation statistics are:
\begin{equation}
\begin{aligned}
    \mathbf S_{t'}^l
    &=(\mathbf X_{t'}^l)^\top\mathbf X_{t'}^l, &
    \mathbf S_{<t}^l
    &=\sum_{t'<t}\mathbf S_{t'}^l .
\end{aligned}
\end{equation}
For any unit direction $\mathbf p$, the quadratic form
$\mathbf p^\top\mathbf S_{<t}^l\mathbf p
=\sum_{t'<t}\|\mathbf X_{t'}^l\mathbf p\|_2^2$ measures its accumulated
activation energy on previous tasks. Large eigenvalues of $\mathbf S_{<t}^l$ indicate repeatedly occupied
visual modes. Since $\mathbf S_{<t}^l$ is positive semi-definite, its SVD
takes the eigendecomposition form:
\begin{equation}
    \mathbf S_{<t}^l
    =
    \mathbf U_{<t}^l
    \boldsymbol{\Lambda}_{<t}^l
    (\mathbf U_{<t}^l)^\top .
\end{equation}
The top-$K_h$ eigenvectors form the historical principal support
$\mathcal H_{<t}^l=\operatorname{span}(\mathbf U_{<t}^{l,K_h})$. This
support may contain transferable visual directions, but rewriting it as a
whole would be too aggressive because many directions encode old
visual-text relations irrelevant to the current task.
\paragraph{Rewritable Shared Mode.}
The shared mode should therefore be a compact portion of the historical
support, selected by its current gradient demand. Under a LoRA update,
when the down-projection direction is fixed as a unit vector
$\mathbf p$, the gradient that can be absorbed by the corresponding
trainable up-projection is proportional to $\mathbf G_t^l\mathbf p$.
Thus, $\|\mathbf G_t^l\mathbf p\|_2^2$ measures the write-in energy
available along $\mathbf p$. We select the historical direction that
maximizes this energy:
\begin{equation}
    \mathbf p_t^{l,S}
    =
    \arg\max_{\substack{
    \mathbf p\in\mathcal H_{<t}^l\\
    \|\mathbf p\|_2=1}}
    \|\mathbf G_t^l\mathbf p\|_2^2 .
\end{equation}
Writing $\mathbf p=\mathbf U_{<t}^{l,K_h}\mathbf q$ with $\|\mathbf q\|_2=1$ turns the objective into $\mathbf q^\top \mathbf M_t^l \mathbf q$, where 
    $\mathbf M_t^l
    =
    (\mathbf U_{<t}^{l,K_h})^\top
    (\mathbf G_t^l)^\top
    \mathbf G_t^l
    \mathbf U_{<t}^{l,K_h}.$
By the Rayleigh-Ritz theorem \cite{horn2012matrix}, the maximizer is the principal eigenvector
of $\mathbf M_t^l$. Therefore, the shared down-projection basis can be defined as:
\begin{equation}
    \mathbf q_t^l=\operatorname{eig}_{\max}(\mathbf M_t^l),
    \qquad
    \mathbf P_t^{l,S}
    =
    \mathbf U_{<t}^{l,K_h}\mathbf q_t^l .
\end{equation}
This basis keeps only the strongest overlap between historical
occupation and current cross-modal demand. $\mathbf P_t^{l,S}$ is the only historical
direction allowed to be rewritten.

\paragraph{Low-Interference Residual Modes.}
After extracting the rewritable shared mode, the remaining gradient demand
is used to construct additional learnable channels for task-specific
adaptation. These residual modes follow the spirit of
interference-avoidance methods that project new updates away from
historical or alignment-sensitive subspaces~\cite{wang2023orthogonal,liang2024inflora,li2025bofa}.
Unlike these methods, our residual modes are not expected to learn the
new task alone. Instead, they provide low-interference plasticity that
cooperates with the shared mode. Concretely, we remove the historical
principal support from the current gradient:
\begin{equation}
    \mathbf G_{t,R}^l
    =
    \mathbf G_t^l
    \big(
    \mathbf I-
    \mathbf U_{<t}^{l,K_h}
    (\mathbf U_{<t}^{l,K_h})^\top
    \big).
\end{equation}
The residual basis is then taken from the dominant right singular
directions of this projected gradient. Specifically, if
$\mathbf G_{t,R}^l=\mathbf U_R^l\boldsymbol{\Sigma}_R^l
(\mathbf V_R^l)^\top$ with
$\mathbf V_R^l=[\mathbf v_{R,1}^l,\ldots,\mathbf v_{R,d_l}^l]$, then we choose:
\begin{equation}
    \mathbf P_t^{l,R}
    =
    [\mathbf v_{R,1}^l,\ldots,\mathbf v_{R,r_R}^l].
\end{equation}
By construction,
$\operatorname{span}(\mathbf P_t^{l,R})\subseteq(\mathcal H_{<t}^l)^\perp$.
Moreover, for any unit residual direction
$\mathbf r\in\operatorname{span}(\mathbf P_t^{l,R})$, the historical
occupation energy satisfies
$\mathbf r^\top\mathbf S_{<t}^l\mathbf r\leq\lambda_{K_h+1}^l$. Residual
modes therefore capture current-task gradient demand outside the 
historical support, while maintaining low overlap with previously
occupied visual modes.
\paragraph{Subspace-Frozen Parameterization.}
The allocation stage only determines where each branch can write. After
allocation, $\mathbf P_t^{l,S}$ and $\mathbf P_t^{l,R}$ are frozen as
the down-projection bases, while only $\mathbf B_t^{l,S}$ and
$\mathbf B_t^{l,R}$ remain trainable. For the first task, without any historical statistics, the bases are
initialized with the dominant right singular vectors of $\mathbf G_1^l$.
The resulting task-$t$ update is
\begin{equation}
    \Delta\mathbf W_t^l
    =
    \mathbf B_t^{l,S}(\mathbf P_t^{l,S})^\top
    +
    \mathbf B_t^{l,R}(\mathbf P_t^{l,R})^\top .
\end{equation}
\subsection{Routed Cross-Modal Training}

Given the allocated bases $\mathbf P_t^{l,S}$ and $\mathbf P_t^{l,R}$,
their up-projections $\mathbf B_t^{l,S}$ and $\mathbf B_t^{l,R}$ are
trained with two routed objectives: a current-task cross-modal objective for new
class learning and an old semantic-structure objective for preserving old
visual-text relations.

\paragraph{Current-Task Cross-Modal Objective.}
For a mini-batch $\mathcal B$, the student model produces row-stacked
normalized image features
$\mathbf Z_t\in\mathbb R^{|\mathcal B|\times d}$. The text embeddings of
current-task classes are
$\mathbf E_t=[\mathbf e_c]_{c\in\mathcal C_t}\in
\mathbb R^{|\mathcal C_t|\times d}$, with batch labels
$\mathbf Y_{\mathcal B}$. The batch image-text logits and current-task
CE loss are:
\begin{equation}
    \mathbf L_t=\tau\mathbf Z_t\mathbf E_t^\top,
    \qquad
    \mathcal L_{\mathrm{c\text{-}ce}}(\mathcal B)
    =
    \operatorname{CE}(\mathbf L_t,\mathbf Y_{\mathcal B}) .
\end{equation}

\paragraph{Old Semantic-Structure Objective.}
The shared basis $\mathbf P_t^{l,S}$ is selected from directions occupied
by old tasks, making it the main channel for reusing transferable
cross-modal knowledge. However, optimizing it may reinterpret these old modes for new-class texts and distort old
visual-text relations. We therefore simply constrain the shared branch using semantic structure induced by old text embeddings. Specifically, before training task $t$, we freeze a copy of the model
obtained after task $t-1$ as the reference teacher. For the same mini-batch
$\mathcal B$, the reference visual features are denoted as
$\bar{\mathbf Z}_{t-1}$, and the old-class text embeddings are collected
as $\mathbf E_{<t}=[\mathbf e_c]_{c\in\mathcal C_{<t}}$. The student and
reference logits over old classes can be represented as:
\begin{equation}
    \mathbf L_t^{<t}
    =
    \tau\mathbf Z_t\mathbf E_{<t}^{\top},
    \qquad
    \bar{\mathbf L}_{t-1}^{<t}
    =
    \tau\bar{\mathbf Z}_{t-1}\mathbf E_{<t}^{\top}.
\end{equation}
The old semantic-structure loss is defined as:

\begin{equation}
\begin{aligned}
\mathcal L_{\mathrm{bi\text{-}kl}}(\mathcal B)
&=
\underbrace{
\tau_C^2
\operatorname{KL}\!\left(
\sigma_C(\bar{\mathbf L}_{t-1}^{<t}/\tau_C)
\,\middle\|\,
\sigma_C(\mathbf L_t^{<t}/\tau_C)
\right)
}_{\text{class-wise structure constraint}}
\\[-1mm]
&\quad+
\underbrace{
\tau_I^2
\operatorname{KL}\!\left(
\sigma_I(\bar{\mathbf L}_{t-1}^{<t}/\tau_I)
\,\middle\|\,
\sigma_I(\mathbf L_t^{<t}/\tau_I)
\right)
}_{\text{instance-wise structure constraint}} .
\end{aligned}
\end{equation}
where $\sigma_C(\cdot)$ and $\sigma_I(\cdot)$ are softmax operators that
normalize over old classes for each image and over batch instances for
each old class-text, respectively. $\tau_C,\tau_I$ are temperatures. The class-wise term
preserves the relative positioning of images with respect to old text
embeddings, while the instance-wise term preserves how each old text
embedding organizes the visual distribution within the batch.

\paragraph{Gradient Routing.}
Gradient routing controls which objective contributes gradients to each
mode. The current-task cross-modal objective sends gradients to both
branches, since both modes participate in new-class learning. In contrast,
the old semantic-structure objective is back-propagated only through the
shared branch, which rewrites directions already occupied by old tasks:
\begin{equation}
\begin{aligned}
    \mathbf B_t^{l,S}
    &\leftarrow
    \mathbf B_t^{l,S}
    -
    \eta\nabla_{\mathbf B_t^{l,S}}
    \Big[
    \mathcal L_{\mathrm{c\text{-}ce}}(\mathcal B)
    +
    \lambda
    \mathcal L_{\mathrm{bi\text{-}kl}}(\mathcal B)
    \Big], \\
    \mathbf B_t^{l,R}
    &\leftarrow
    \mathbf B_t^{l,R}
    -
    \eta\nabla_{\mathbf B_t^{l,R}}
    \mathcal L_{\mathrm{c\text{-}ce}}(\mathcal B).
\end{aligned}
\end{equation}
Here $\eta$ is the learning rate and $\lambda$ controls the strength of
old-structure constraint. As a result, the shared mode learns
transferable alignment under old semantic constraints, while the residual
mode captures task-specific evidence through low-interference
directions.

\subsection{Reliability-Guided Bridge-Prototype Ensemble}

Text embeddings offer stable semantic anchors, but they
mainly encode high-level semantics and may miss discriminative evidence
from the visual side. Existing methods often combine text embeddings and
visual prototypes for prediction~\cite{zhou2025engine,li2025bofa,he2025harnessing},
but such fusion is usually coarse: the two modal representations are
treated as fixed endpoints, and a global fusion rule cannot reflect that
different classes may rely on different degrees of textual semantics and
visual evidence. We therefore search the \emph{inter-modal} discriminative
space on the unit hypersphere by constructing class-wise geodesic bridges
and estimating which bridge depths are reliable. For each class $c$ introduced at task $t$, we define its sample set as
$\mathcal X_c^t=\{x_i \mid (x_i,y_i)\in\mathcal D_t,\ y_i=c\}$. Its
normalized visual prototype is computed as:
\begin{equation}
    \mathbf v_c
    =
    \frac{
    \sum_{x_i\in\mathcal X_c^t}\mathbf z_i}
    {\left\|
    \sum_{x_i\in\mathcal X_c^t}\mathbf z_i
    \right\|_2}.
\end{equation}
Given $\mathbf v_c$ and the normalized text embedding $\mathbf e_c$, the
bridge depth is discretized as
$\{\alpha_1, \alpha_2,\ldots, \alpha_{K_b}\}$, and the endpoint angle on the unit
hypersphere is
$\theta_c=\arccos(\mathbf v_c^\top\mathbf e_c)$. The bridge prototype at
depth $\alpha_k$ is obtained by geodesic interpolation:
\begin{equation}
    \mathbf b_c(\alpha_k)
    =
    \frac{
    \sin((1-\alpha_k)\theta_c)\mathbf v_c
    +
    \sin(\alpha_k\theta_c)\mathbf e_c}
    {\sin\theta_c}.
\end{equation}
Here $\alpha_k=0$ and $\alpha_k=1$ correspond to the visual and textual
endpoints, respectively. Compared with linear interpolation~\cite{li2025bofa}, geodesic
interpolation provides a smooth transition on the unit hypersphere from
the visual semantic coordinate to the textual one, so bridge prototypes at
each depth remain semantically comparable across classes despite the
modality gap. For a fixed depth $\alpha_k$, we instantiate a bridge classifier using
the bridge prototypes of all seen classes. Its logits for class
$j\in\mathcal C_{\leq t}$ on image $x_i$ is
$\tau\mathbf z_i^\top\mathbf b_j(\alpha_k)$, and the corresponding
softmax probability for class $c$ is denoted by $p_{\alpha_k}(c|x_i)$.
The reliability of depth $\alpha_k$ for class $c$ is then measured by the
average correct-class confidence on these samples as
    $\rho_c(\alpha_k)
    =
    \frac{1}{|\mathcal X_c^t|}
    \sum_{x_i\in\mathcal X_c^t}
    p_{\alpha_k}(c|x_i).$
The reliability weights are then normalized over all bridge depths by:
\begin{equation}
    \pi_c(\alpha_k)
    =
    \frac{\exp(\rho_c(\alpha_k)/\beta)}
    {\sum_{m=1}^{K_b}\exp(\rho_c(\alpha_m)/\beta)} .
\end{equation}
The temperature $\beta$ controls how sharply the ensemble concentrates
on highly reliable depths. For old classes, the visual prototypes and
reliability weights are stored when the classes are first introduced, and
no raw old data is kept.

At inference, prediction is made by a reliability-weighted ensemble of
bridge-prototype logits:
\begin{equation}
    \hat y
    =
    \arg\max_{c\in\mathcal C_{\leq t}}
    \sum_{k=1}^{K_b}
    \pi_c(\alpha_k)
    \tau\mathbf z(x)^\top
    \mathbf b_c(\alpha_k).
\end{equation}
This ensemble improves CIL classification by exploiting discriminative bridge
prototypes in the inter-modal semantic region. Its effectiveness relies on
a stably evolving inter-modal semantic structure. As shown in
Table~\ref{tab:ablation_bridge}, methods without explicit
semantic-structure preservation do not obtain consistent gains from the
bridge-prototype ensemble. Thus, the classifier is tightly coupled with
the proposed dual-mode low-rank learner. Further evidence and analysis of the bridge-prototype ensemble are
provided in Appendix~A.3.

\section{Experiments}

\subsection{Experimental Setup}
\paragraph{Datasets.}
We evaluate DuLBE under two standard CLIP-based CIL settings adopted in
recent studies. \textbf{Setting-A}~\cite{huang2024rapf} uses OpenAI
CLIP~\cite{radford2021learning} and includes
CIFAR~\cite{krizhevsky2009learning}, CUB~\cite{wah2011caltech},
ImageNet-R~\cite{hendrycks2021many}, and
ImageNet100~\cite{deng2009imagenet}. \textbf{Setting-B}~\cite{zhou2025engine}
uses OpenCLIP~\cite{cherti2023reproducible}
pretrained on LAION-400M~\cite{schuhmann2021laion} and we include CIFAR,
SUN~\cite{xiao2010sun}, Cars~\cite{krause2013cars}, and
Food~\cite{bossard2014food}. For each setting, we use the same task
splits and data preprocessing as the corresponding protocol.

\paragraph{Implementation Details.}
All experiments are implemented in PyTorch and conducted on an NVIDIA
RTX 4090 GPU. For fair comparison, all methods use the same ViT-B/16
pretrained weights required by each setting: OpenAI CLIP in Setting-A and
LAION-400M-pretrained OpenCLIP in Setting-B. The proposed dual-mode low-rank learner
is inserted into the transformer blocks of CLIP's visual tower, including
the key/value attention projections and MLP layers. The historical
support dimension is $K_h=128$, with shared rank $r_S=1$ and residual
rank $r_R=8$. The KL temperatures
are set to $\tau_C=5$ and $\tau_I=0.1$, respectively, and the
semantic-structure loss weight is $\lambda=0.5$. We optimize the model
with Adam~\cite{kingma2015adam} using a learning rate of $1\times10^{-3}$
and a batch size of 32. For the bridge-prototype
ensemble, we sample $K_b=10$ bridge depths and set the reliability
temperature to $\beta=0.05$.

\subsection{Comparison with the State-of-the-Arts}


\begin{table*}[!t]
\centering

\begingroup
\small
\setlength{\tabcolsep}{2.4pt}
\renewcommand{\arraystretch}{0.95}

\begin{tabular*}{\textwidth}
{@{\extracolsep{\fill}}lcccccccccccccccc@{}}
\toprule
\multicolumn{1}{c}{Method}
& \multicolumn{8}{c}{Setting-A~(CLIP)}
& \multicolumn{8}{c}{Setting-B~(OpenCLIP)} \\
\cmidrule(lr){2-9}\cmidrule(lr){10-17}

& \multicolumn{2}{c}{CIFAR}
& \multicolumn{2}{c}{CUB}
& \multicolumn{2}{c}{I.N.-R}
& \multicolumn{2}{c}{I.N.100}
& \multicolumn{2}{c}{CIFAR}
& \multicolumn{2}{c}{SUN}
& \multicolumn{2}{c}{Cars}
& \multicolumn{2}{c}{Food} \\

& $\bar{\mathcal{A}}$ & $\mathcal{A}_l$
& $\bar{\mathcal{A}}$ & $\mathcal{A}_l$
& $\bar{\mathcal{A}}$ & $\mathcal{A}_l$
& $\bar{\mathcal{A}}$ & $\mathcal{A}_l$
& $\bar{\mathcal{A}}$ & $\mathcal{A}_l$
& $\bar{\mathcal{A}}$ & $\mathcal{A}_l$
& $\bar{\mathcal{A}}$ & $\mathcal{A}_l$
& $\bar{\mathcal{A}}$ & $\mathcal{A}_l$ \\
\cmidrule[\lightrulewidth](r){1-9}%
\cmidrule[\lightrulewidth](l){10-17}

ContinualCLIP
& -- & 66.7 & -- & 51.2 & -- & 72.0 & -- & 75.4
& -- & 71.4 & -- & 72.1 & -- & 76.4 & -- & 81.9 \\

DualPrompt~(ECCV'22)
& 81.5 & 72.5 & -- & -- & 82.0 & 75.8 & 80.7 & 67.4
& 81.6 & 72.4 & 82.5 & 74.4 & 76.3 & 62.9 & 84.9 & 77.3 \\

CODA~(CVPR'23)
& 77.0 & 62.3 & 66.6 & 50.9 & 78.0 & 67.5 & 64.1 & 34.8
& 82.4 & 73.4 & 83.3 & 75.7 & 80.2 & 66.5 & 86.2 & 78.8 \\
\cmidrule[\lightrulewidth](r){1-9}%
\cmidrule[\lightrulewidth](l){10-17}

PROOF~(TPAMI'25)
& 86.2 & 76.3 & -- & -- & 82.8 & 77.1 & 84.7 & 72.5
& 86.8 & 79.1 & 83.9 & 77.3 & 90.7 & 86.5 & 90.0 & 84.7 \\

RAPF~(ECCV'24)
& 86.2 & 79.0 & 82.7 & 76.2 & 85.6 & 80.3
& \underline{87.5} & \underline{80.2}
& 86.1 & 78.0 & 82.1 & 72.5 & 82.9 & 62.9 & 88.6 & 81.2 \\

CLG-CBM~(CVPR'25)
& 84.9 & 76.9 & 82.9 & 77.8 & -- & -- & 86.0 & 78.5
& -- & -- & -- & -- & -- & -- & -- & -- \\

ENGINE~(ICCV'25)
& 82.1 & 73.1 & 83.9 & 76.2 & 84.4 & 77.0 & -- & --
& 86.9 & 79.2 & 85.0 & 78.5
& \underline{94.1} & 90.1
& \underline{89.8} & \underline{83.9} \\

DesCLIP~(TMM'26)
& 85.6 & 78.7
& \underline{85.1} & \underline{78.7}
& 86.4 & 77.8 & 80.7 & 72.2
& -- & -- & -- & -- & -- & -- & -- & -- \\

MG-CLIP~(ICCV'25)
& \underline{87.0} & 80.6
& 80.6 & 72.0
& 87.6 & \underline{82.7}
& 87.3 & 78.4
& \underline{89.1} & \underline{82.4}
& 55.9 & 43.0
& 88.4 & 80.7
& 88.2 & 82.2 \\

BOFA~(AAAI'26)
& -- & -- & -- & -- & -- & -- & -- & --
& 86.5 & 79.3
& \underline{85.6} & \underline{78.9}
& 93.8 & 89.2
& 89.0 & 82.7 \\

LoDA-CLIP~(ICML'26)
& 86.8 & \underline{81.5}
& 84.2 & 78.1
& \underline{87.8} & 82.0
& 85.9 & 76.4
& 88.9 & 81.7
& -- & --
& \textbf{94.3} & \textbf{90.6}
& -- & -- \\
\cmidrule[\lightrulewidth](r){1-9}%
\cmidrule[\lightrulewidth](l){10-17}

\textbf{DuLBE}~(Ours)
& \textbf{89.5} & \textbf{84.2}
& \textbf{86.1} & \textbf{80.6}
& \textbf{88.4} & \textbf{84.3}
& \textbf{88.3} & \textbf{80.7}
& \textbf{90.0} & \textbf{84.5}
& \textbf{86.7} & \textbf{80.3}
& \underline{94.1} & \underline{90.3}
& \textbf{91.3} & \textbf{86.7} \\

\bottomrule
\end{tabular*}
\endgroup

\caption{Main comparison under two standard CLIP-based CIL settings.
Setting-A~\cite{huang2024rapf} uses OpenAI CLIP ViT-B/16, and
Setting-B~\cite{zhou2025engine} uses OpenCLIP ViT-B/16 pretrained on
LAION-400M. All methods are evaluated using the standard 10-task split without \emph{external training data}. The
best results are highlighted in \textbf{bold}, and the second-best results
are \underline{underlined}.}
\label{tab:main_results}
\end{table*}

We compare DuLBE with representative baselines, including the zero-shot
ContinualCLIP baseline~\cite{thengane2022clip}, prompt-based methods DualPrompt~\cite{wang2022dualprompt} and
CODA~\cite{smith2023coda}, and recent CLIP-based CIL methods
PROOF~\cite{zhou2025proof}, RAPF~\cite{huang2024rapf}, SGCL~\cite{yu2024exploiting}, CLG-CBM~\cite{yu2025language},
ENGINE~\cite{zhou2025engine}, DesCLIP~\cite{he2026desclip},
MG-CLIP~\cite{huang2025mind}, BOFA~\cite{li2025bofa}, LoDA-CLIP~\cite{he2026loda}, and LfI~\cite{gong2026learning}. We report the average accuracy over all incremental
stages, denoted as $\bar{\mathcal A}$, and the final accuracy after the
last task, denoted as $\mathcal A_l$.

As shown in Table~\ref{tab:main_results}, DuLBE achieves the strongest
overall performance under both standard settings. In Setting-A, DuLBE
ranks first on all four datasets and all reported metrics. The advantage
is particularly evident on CIFAR, where DuLBE improves
$\bar{\mathcal A}$ and $\mathcal A_l$ over the best previous results by
2.5\% and 2.7\%, respectively. It also obtains clear final-accuracy gains
on CUB and ImageNet-R, demonstrating that the proposed dual-mode
allocation benefits both general object recognition and fine-grained
visual adaptation. In Setting-B, DuLBE achieves the best results on most
datasets and shows especially strong retention on CIFAR and Food, with
$\mathcal A_l$ improvements of 2.1\% and 2.0\% over the second. Broader baseline comparisons and backbone scalability
results are reported in Appendix~C.1 and C.3, respectively.

\begin{figure}
    \centering
    \includegraphics[width=\linewidth]{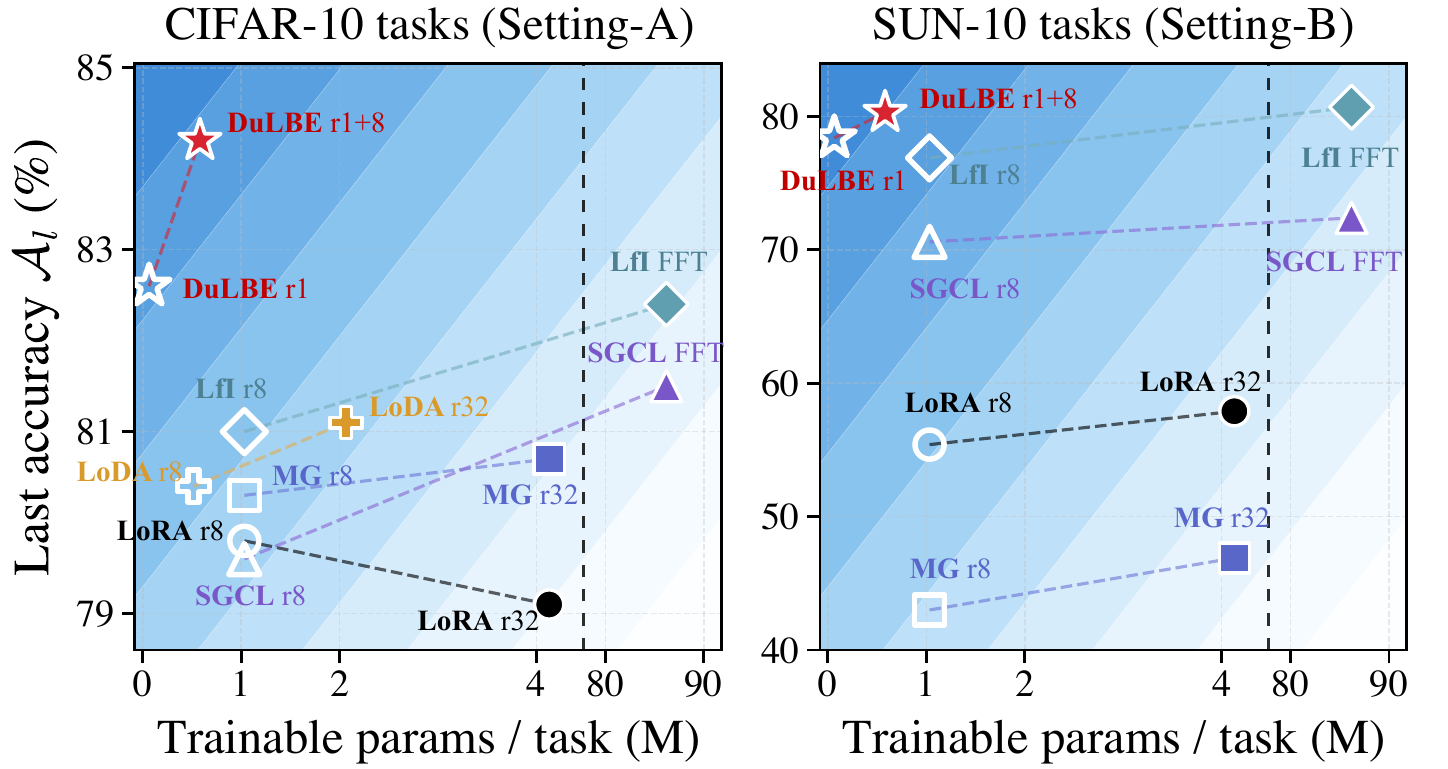}
\caption{Efficiency-performance comparison with benchmark methods. For a
fair comparison, trainable parameters are counted only from the key/value
attention projections and MLP layers in CLIP's visual tower. Here, $r$
denotes the low-rank dimension, and FFT denotes full fine-tuning.}
\label{fig:detail_comparison}
\end{figure}

We further provide a detailed \textit{efficiency-performance} comparison in
Figure~\ref{fig:detail_comparison}. Compared with parameter-tuning methods: LoRA-CLIP,
MG-CLIP, LfI, LoDA-CLIP, and SGCL under different adaptation budgets,
DuLBE achieves a more favorable trade-off. In particular, the \textit{shared-mode-only}
variant DuLBE$_{r1}$ uses only about 0.065M trainable parameters in the visual tower,
while DuLBE$_{r1+8}$ still requires only about 0.58M parameters after
adding the \textit{residual-mode} low-rank learner. Despite these modest parameter budgets, DuLBE
remains in the upper-left region of the plot, outperforming higher-rank LoRA variants and full fine-tuning counterparts.
This confirms that the improvement of DuLBE does not arise from increasing
the adaptation scale, but from allocating the key subspaces for
transferable shared knowledge and low-interference task-specific
variations. Moreover, we provide detailed analyses of training, inference, and CIL
memory overheads in Appendix~C.5.

\begin{figure*}[!t]
    \centering
    \includegraphics[width=0.9\linewidth]{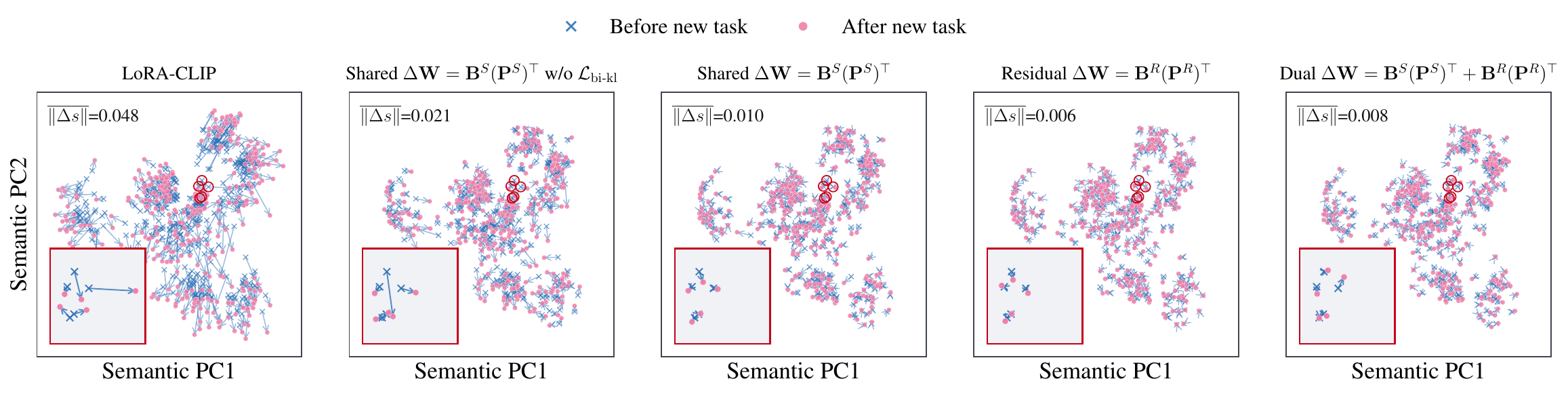}
\caption{Visualization of old-task semantic-coordinate drift. Each old sample is represented by its cosine-similarity vector to Task-1 class-text embeddings and projected to two dimensions. Arrows show the shift after learning the new task, and $\overline{\Vert\Delta s\Vert}$ denotes the average drift in the original text-semantic coordinate space.}
\label{fig:vis}
\end{figure*}

\subsection{Further Analysis}
\label{sec:further_analysis}

\begin{table}[t]
\centering
\small
\setlength{\tabcolsep}{2.2pt}
\renewcommand{\arraystretch}{0.8}
\begin{tabular*}{\columnwidth}{@{\extracolsep{\fill}}lcccccccc@{}}
\toprule
\multicolumn{1}{l}{Mode bases}
& \multicolumn{2}{c}{$\nabla_{\mathcal L_{\mathrm{c\text{-}ce}}}$}
& \multicolumn{2}{c}{$\nabla_{\mathcal L_{\mathrm{bi\text{-}kl}}}$}
& \multicolumn{2}{c}{CIFAR (A)}
& \multicolumn{2}{c}{CUB (A)} \\
\cmidrule(lr){2-3}
\cmidrule(lr){4-5}
& $\mathbf B^{S}$ & $\mathbf B^{R}$
& $\mathbf B^{S}$ & $\mathbf B^{R}$
& $\bar{\mathcal A}\uparrow$ & $\mathcal A_l\uparrow$
& $\bar{\mathcal A}\uparrow$ & $\mathcal A_l\uparrow$ \\
\midrule
LoRA$^{*}$ ($r=32$)
& -- & -- & -- & --
& (86.1) & (78.5) & (81.0) & (73.8) \\
\midrule
Shared $\mathbf P^{S}$
& \checkmark & -- & \checkmark & $\times$
& +2.4 & +4.1 & +3.2 & +4.0 \\
Residual $\mathbf P^{R}$
& -- & \checkmark & $\times$ & $\times$
& +2.5 & +4.4 & +4.7 & +5.5 \\
Dual $[\mathbf P^{S},\mathbf P^{R}]$
& \checkmark & \checkmark & $\times$ & $\times$
& +1.3 & +2.4 & +1.1 & +1.5 \\
Dual $[\mathbf P^{S},\mathbf P^{R}]$
& \checkmark & \checkmark & \checkmark & \checkmark
& +2.6 & +4.4 & +4.2 & +5.3 \\
Dual $[\mathbf P^{S},\mathbf P^{R}]$
& \checkmark & \checkmark & \checkmark & $\times$
& \textbf{+3.4} & \textbf{+5.7}
& \textbf{+5.1} & \textbf{+6.8} \\
\bottomrule
\end{tabular*}
\caption{Ablation of dual-mode collaborative learning. All values are
improvements over the LoRA$^{*}$ baseline, which replaces the low-rank learner with standard LoRA while keeping all other
components unchanged.}
\label{tab:ablation_dualmode}
\end{table}

\paragraph{Effectiveness of Dual-Mode Collaborative Learning.}
Table~\ref{tab:ablation_dualmode} shows that both shared and residual
modes improve over the LoRA baseline through complementary roles. The
shared mode reuses historically occupied directions but remains sensitive
to unconstrained rewriting, while the residual mode captures task-specific
variations with lower interference. This distinction is supported by
Figure~\ref{fig:vis}, where semantic-structure constraint substantially
reduces the drift of the shared mode, whereas the residual mode naturally
maintains low drift. Simply combining both modes provides limited gains,
indicating that mode allocation alone is insufficient for effective
collaboration. Regularizing both modes, however, restricts residual
plasticity. Routing $\mathcal L_{\mathrm{bi\text{-}kl}}$ only to the shared
mode enables controlled knowledge reuse while preserving flexible residual
adaptation, yielding the greatest gains of 3.4/5.7\% on CIFAR and 5.1/6.8\% on CUB in
$\bar{\mathcal A}/\mathcal A_l$.

\paragraph{Analysis of Bridge-Prototype Ensemble.}
Table~\ref{tab:ablation_bridge} evaluates the bridge-prototype ensemble.
Fixed bridge depths (rows 1--3) consistently outperform the text-only classifier,
confirming the discriminative value of the inter-modal region. Uniformly
combining multiple depths remains effective but is not consistently
superior to the best fixed depth, as different classes may favor
different bridge regions. Reliability weighting captures this variation
and achieves the best performance on both datasets. However, its effect on other
learners is less consistent: it improves LoRA-CLIP but degrades LfI and
RAPF. Thus, the ensemble is not a universally effective plug-in. Its
benefit depends on a compatible inter-modal structure, which DuLBE
explicitly stabilizes.

\begin{table}[t]
\centering
\small
\setlength{\tabcolsep}{0.8pt}
\renewcommand{\arraystretch}{0.8}
\begin{tabular*}{\columnwidth}{@{\extracolsep{\fill}}lcccccc@{}}
\toprule
Method
& \multicolumn{2}{c}{Text Clf.}
& \multicolumn{4}{c}{B.E. Clf.} \\
\cmidrule(lr){2-3}
\cmidrule(l){4-7}
& CUB & Food
& $\alpha$ & Rel.
& CUB & Food \\
\midrule
Ours
& 67.9 & 83.7 & 0.25 & $\times$ & 77.9 & 85.0 \\
Ours
& -- & -- & 0.50 & $\times$ & 77.7 & 83.9 \\
Ours
& -- & -- & 0.75 & $\times$ & 72.9 & 84.2 \\
Ours
& -- & -- & $\{\alpha_k\}$ & $\times$ & 79.1 & 84.6 \\
Ours
& -- & -- & $\{\alpha_k\}$ & \checkmark
& \textbf{80.6}(+12.7) & \textbf{86.7}(+3.0) \\
\midrule
LoRA-CLIP$_{r8}$
& 63.2 & 79.6 & $\{\alpha_k\}$ & \checkmark
& 73.6(+10.4) & 80.0(+0.4) \\
RAPF
& 76.2 & 81.2 & $\{\alpha_k\}$ & \checkmark
& 71.9(-4.3) & 74.2(-7.0) \\
LfI$_{r8}$
& 79.3 & 87.1 & $\{\alpha_k\}$ & \checkmark
& 75.4(-3.9) & 84.0(-3.1) \\
\bottomrule
\end{tabular*}
\caption{Ablation of the bridge-prototype ensemble on CUB (A) and Food
(B). All entries report the last accuracy $\mathcal A_l$. B.E. denotes
bridge-prototype ensemble and Rel. denotes reliability weighting over
bridge depths.}
\label{tab:ablation_bridge}
\end{table}

\begin{figure}[!t]
    \centering
    \includegraphics[width=\linewidth]{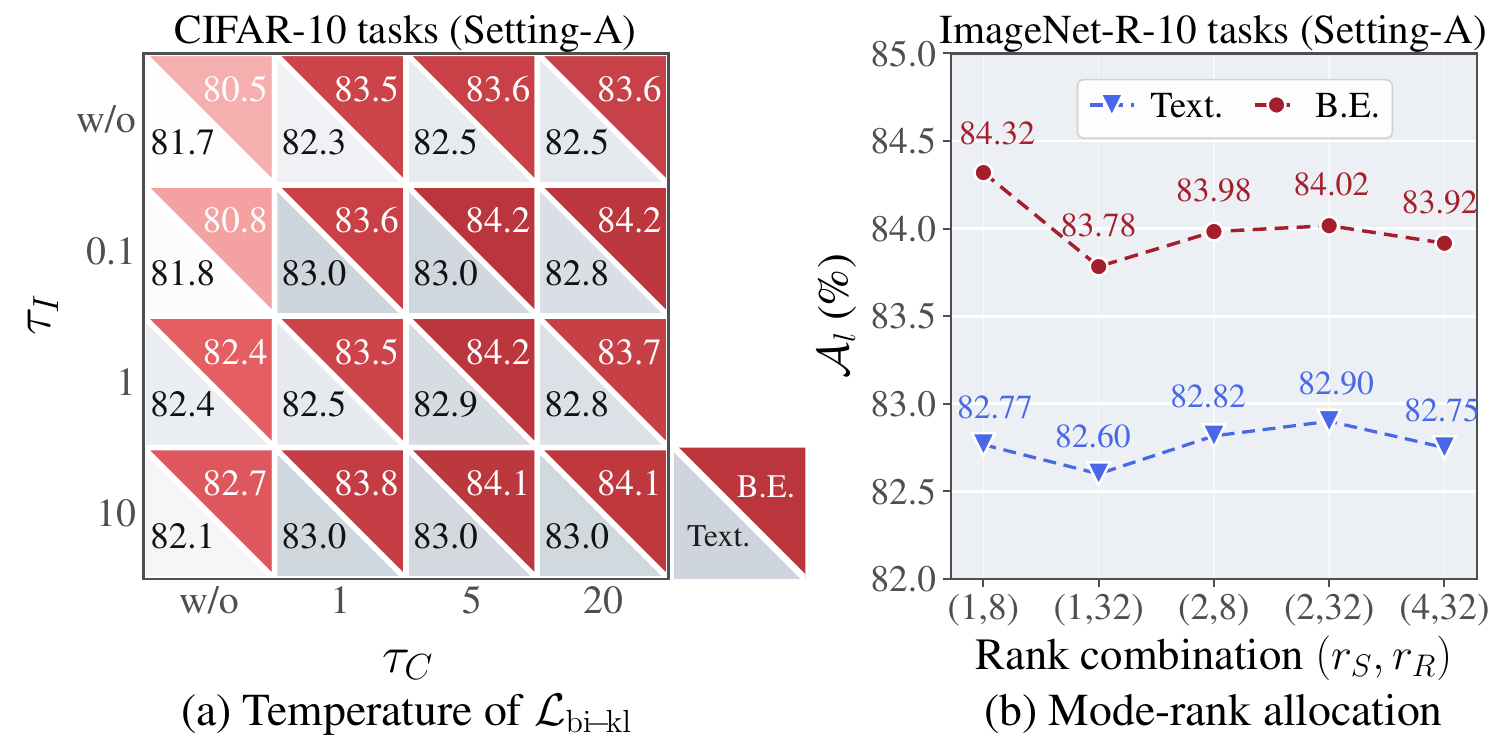}
\caption{Hyper-parameter analysis of (a) the class-wise and instance-wise
temperatures in $\mathcal L_{\mathrm{bi\text{-}kl}}$ and (b) the
shared/residual rank allocation.}
\label{fig:hyper-params}
\end{figure}

\paragraph{Hyper-parameter Analysis.}
DuLBE is generally robust to the KL temperatures and mode-rank allocation. As shown in Figure~\ref{fig:hyper-params}, the setting $\tau_C=5$ and $\tau_I=0.1$ achieves the best bridge-ensemble accuracy, while nearby choices yield comparable results. Moreover, the compact rank allocation $(r_S,r_R)=(1,8)$ performs best, and increasing the ranks brings no consistent benefit. This suggests that a single compact direction is sufficient for knowledge transfer, while a modest residual space provides adequate plasticity. Additionally, the number of geodesic bridge depths $K_b$ and the reliability-smoothing temperature $\beta$ are further analyzed in
Appendix~C.4.

\section{Conclusion}

We presented DuLBE, an exemplar-free Vision-Language CIL framework that
addresses continual adaptation with a dual-mode low-rank learner and a
bridge-prototype ensemble classifier. Through subspace allocation and routed objective learning, DuLBE allocates transferable
cross-modal knowledge to a compact rewritable shared mode and
task-specific variations to low-interference residual modes, enabling
controlled knowledge reuse and flexible adaptation. Building on the
resulting stable inter-modal structure, the bridge-prototype ensemble
classifier exploits discriminative prototypes in the inter-modal semantic region
with bridge-depth reliability. Experiments under standard CIL settings show that DuLBE
achieves state-of-the-art performance while preserving
the parameter efficiency of low-rank adaptation.

\newpage

\clearpage
\bibliography{references}

\begin{bibunit}[aaai2027]
\def\DuLBEAppendixEmbedded{1}
\graphicspath{{supplement_source/}}
\ifdefined\DuLBEAppendixEmbedded
\else
\documentclass[letterpaper]{article} 
\usepackage[preprint]{aaai2027}  
\usepackage[hyphens]{url}  
\usepackage{graphicx} 
\usepackage{pifont}
\urlstyle{rm} 
\def\UrlFont{\rm}  
\usepackage{natbib}  
\usepackage{caption} 
\frenchspacing  

\usepackage[ruled,vlined,linesnumbered,algo2e]{algorithm2e}
\usepackage{framed}
\usepackage{float}
\usepackage{amsmath}
\usepackage{amssymb}
\usepackage{booktabs}
\usepackage{array}
\usepackage{tabularx}
\usepackage{multirow}

\definecolor{DuLBEGray}{RGB}{242,243,245}
\definecolor{DuLBEBlue}{RGB}{229,239,250}
\definecolor{DuLBEYellow}{RGB}{252,246,220}
\definecolor{DuLBEOrange}{RGB}{180,70,0}

\makeatletter
\newcommand{\DuLBEModuleTitle}[1]{%
    \algocf@seteveryparnl{\relax}%
    \textbf{#1}\par
    \algocf@linesnumbered}
\makeatother

\pdfinfo{
/TemplateVersion (2027.1)
}

\title{Dual-Mode Low-Rank Learner with
Bridge-Prototype Ensemble for Vision-Language Class-Incremental Learning\\
{\Large Appendix}}

\author{
    Chiyuan He\textsuperscript{\rm 1},
    Zihuan Qiu\textsuperscript{\rm 1},
    Fanman Meng\textsuperscript{\rm 1,\ding{41}},
    Chao Wang\textsuperscript{\rm 2},
    Liangjiang Chen\textsuperscript{\rm 1},
    Linfeng Xu\textsuperscript{\rm 1},
    Qingbo Wu\textsuperscript{\rm 1},
    Hongliang Li\textsuperscript{\rm 1}
}
\affiliations{
    \textsuperscript{\rm 1}University of Electronic Science and Technology of China, Chengdu, China\\
    \textsuperscript{\rm 2}Qiyuan Lab, Beijing, China\\
    \{cyhe,zihuanqiu,202522011613\}@std.uestc.edu.cn\\
    \{lfxu,qbwu,hlli\}@uestc.edu.cn,\quad
    w-c15@tsinghua.org.cn
}

\copyrighttext{\ding{41}\ Corresponding author: fmmeng@uestc.edu.cn.\quad
Preprint. Work in progress.}
\fi

\setcounter{secnumdepth}{2}
\setcounter{section}{0}
\setcounter{table}{0}
\setcounter{figure}{0}
\setcounter{equation}{0}
\setcounter{algocf}{0}
\renewcommand{\thesection}{\Alph{section}}
\renewcommand{\thetable}{\Alph{section}.\arabic{table}}
\renewcommand{\thefigure}{\Alph{section}.\arabic{figure}}
\renewcommand{\theequation}{\Alph{section}.\arabic{equation}}
\renewcommand{\thealgocf}{\Alph{section}.\arabic{algocf}}
\numberwithin{table}{section}
\numberwithin{figure}{section}
\numberwithin{equation}{section}
\numberwithin{algocf}{section}

\ifdefined\DuLBEAppendixEmbedded
\DuLBEAppendixTitle
\else
\begin{document}
\maketitle
\fi
\raggedbottom

\section*{Overview}
The appendix is organized into four parts.
Section~\ref{sec:method_analysis} analyzes the main design choices and
their supporting evidence. Section~\ref{sec:experimental_setup} details
the experimental protocols and training configurations.
Section~\ref{sec:additional_results} reports additional experimental results. Section~\ref{sec:complete_procedure} gives the
complete procedure of proposed DuLBE.

\section{Method Analysis and Evidence}
\label{sec:method_analysis}

\subsection{Rewritable Shared Mode}

This subsection supplements the rewritable \emph{Shared Mode} in the
main paper by explaining why DuLBE selects a compact, layer-wise
direction instead of making a large-rank historical subspace jointly
rewritable. For each adapted layer $l$, the accumulated historical
statistic is decomposed as
\begin{equation}
    \mathbf S_{<t}^l
    =
    \mathbf U_{<t}^l
    \boldsymbol\Lambda_{<t}^l
    (\mathbf U_{<t}^l)^\top.
\end{equation}
The top-$K_h$ eigenvectors
$\mathbf U_{<t}^{l,K_h}$ define the historical principal support
$\mathcal H_{<t}^l$. Directions in this support have been repeatedly
activated by previous tasks, but only a small subset may also be demanded
by current task. Therefore, consider a rank-$r$ shared basis
$\mathbf P_t^{l,S}(r)\subseteq \operatorname{span}(\mathcal H_{<t}^l)$. With the basis fixed,
one gradient step on its trainable up-projection gives
\begin{equation}
\begin{aligned}
    &\mathcal L_{\mathrm{c\text{-}ce}}
    (\mathbf W_{t-1}^l+\Delta\mathbf W_t^{l,S})
    -
    \mathcal L_{\mathrm{c\text{-}ce}}
    (\mathbf W_{t-1}^l) \\
    &\hspace{12mm}
    =
    -\eta
    \|\mathbf G_t^l\mathbf P_t^{l,S}(r)\|_F^2
    +O(\eta^2).
\end{aligned}
\label{eq:layer_shared_descent}
\end{equation}
Thus, the projected gradient energy measures how much current-task
learning signal can be absorbed through the selected historical
directions. To maximize this energy, define
\begin{equation}
    \mathbf M_t^l
    =
    (\mathbf U_{<t}^{l,K_h})^\top
    (\mathbf G_t^l)^\top\mathbf G_t^l
    \mathbf U_{<t}^{l,K_h},
\end{equation}
and let $\mu_1^l\geq\cdots\geq\mu_{K_h}^l$ be its eigenvalues, with
corresponding eigenvectors $\{\mathbf q_i^l\}$. By the Ky Fan maximum
principle~\cite{horn2012matrix}, the optimal rank-$r$ shared basis is
\begin{equation}
\label{eq:layer_shared_basis}
\begin{gathered}
\mathbf P_t^{l,S}(r)
=
\mathbf U_{<t}^{l,K_h}
[\mathbf q_1^l,\ldots,\mathbf q_r^l], \\[1mm]
\left\|
\mathbf G_t^l\mathbf P_t^{l,S}(r)
\right\|_F^2
=
\sum_{i=1}^{r}\mu_i^l .
\end{gathered}
\end{equation}
Therefore, $\mu_i^l$ represents the marginal current-task benefit of
making the corresponding historical direction rewritable.

Although a larger shared rank captures more gradient energy, it also
opens more historically occupied directions to modification. From a
sparse-update perspective, this trade-off can be expressed by the
analytical surrogate
\begin{equation}
    \max_{1\leq r\leq K_h}
    \left[
    \sum_{i=1}^{r}\mu_i^l-\gamma_l r
    \right],
    \label{eq:sparse_shared_tradeoff}
\end{equation}
where $\gamma_l$ denotes the conceptual cost of exposing one additional
historical direction to rewriting and is not an extra training
hyperparameter. Increasing the shared rank from $r$ to $r+1$ is useful
only when its additional gradient benefit
$\mu_{r+1}^l$ exceeds this rewriting cost. When the projected spectrum is
concentrated, the leading direction retains most transferable gradient
energy, whereas the remaining directions provide limited benefit while
increasing the risk of disturbing old visual-text relations. DuLBE
therefore adopts the sparse choice $r_S=1$:
\begin{equation}
    \mathbf P_t^{l,S}
    =
    \mathbf U_{<t}^{l,K_h}
    \operatorname{eig}_{\max}(\mathbf M_t^l).
\end{equation}
This layer-wise construction retains the strongest overlap between
historical occupation and current cross-modal demand while avoiding
unnecessary high-rank rewriting.

\subsection{Residual Modes and Gradient Routing}

This subsection supplements the low-interference \emph{Residual Modes}
and \emph{Gradient Routing} in the main paper. We analyze two connected
aspects: how the residual mode captures task-specific information not
covered by the compact shared mode, and why routing the KL-constraint
gradient to this space would introduce noise into its plastic
exploration.

\paragraph{Residual-mode construction.}
Because $\mathbf P_t^{l,S}$ already represents the most useful direction
inside $\mathcal H_{<t}^l$, allocating additional residual capacity in
the same support would duplicate the shared branch and expose more
historical structure to modification. DuLBE therefore projects the
prospective gradient onto the complementary space:
\begin{equation}
\begin{aligned}
    \boldsymbol\Pi_{<t}^{l,\perp}
    &=
    \mathbf I-
    \mathbf U_{<t}^{l,K_h}
    (\mathbf U_{<t}^{l,K_h})^\top, \\
    \mathbf G_{t,R}^l
    &=
    \mathbf G_t^l
    \boldsymbol\Pi_{<t}^{l,\perp}.
\end{aligned}
\label{eq:residual_projected_gradient}
\end{equation}
This removes gradient components associated with the principal
historical support while retaining current-task demand that cannot be
absorbed by the shared mode. Let
\begin{equation}
    \mathbf G_{t,R}^l
    =
    \mathbf A_t^l
    \boldsymbol\Sigma_t^l
    (\mathbf V_t^l)^\top
\end{equation}
be its singular value decomposition, with singular values
$\sigma_1^l\geq\sigma_2^l\geq\cdots$. We choose to select
\begin{equation}
    \mathbf P_t^{l,R}
    =
    [\mathbf v_{t,1}^l,\ldots,\mathbf v_{t,r_R}^l].
\end{equation}
For any orthonormal basis $\mathbf R$ satisfying
$\mathbf R^\top\mathbf U_{<t}^{l,K_h}=\mathbf0$, we have
$\mathbf G_t^l\mathbf R=\mathbf G_{t,R}^l\mathbf R$. The variational
characterization of singular values therefore gives
\begin{equation}
\label{eq:optimal_residual_basis}
\begin{gathered}
\mathbf P_t^{l,R}
\in
\underset{\substack{
    \mathbf R^\top\mathbf R=\mathbf I_{r_R}\\
    \mathbf R^\top\mathbf U_{<t}^{l,K_h}=\mathbf 0
}}{\arg\max}
\left\|\mathbf G_t^l\mathbf R\right\|_F^2, \\[1mm]
\left\|\mathbf G_t^l\mathbf P_t^{l,R}\right\|_F^2
=
\sum_{i=1}^{r_R}(\sigma_i^l)^2 .
\end{gathered}
\end{equation}
Accordingly, the residual basis is not an arbitrary orthogonal
component: it is the rank-$r_R$ subspace outside the historical support
that retains the largest current-task gradient energy. A gradient step
through this basis provides the first-order loss decrease:
\begin{equation}
\begin{aligned}
    &\mathcal L_{\mathrm{c\text{-}ce}}
    (\mathbf W_{t-1}^l+\Delta\mathbf W_t^{l,R})
    -
    \mathcal L_{\mathrm{c\text{-}ce}}
    (\mathbf W_{t-1}^l) \\
    &\hspace{14mm}
    =
    -\eta
    \sum_{i=1}^{r_R}(\sigma_i^l)^2
    +O(\eta^2).
\end{aligned}
\label{eq:residual_first_order_descent}
\end{equation}
The same construction also limits historical interference. Let
$\rho_{K_h+1}^l$ denote the largest eigenvalue of
$\mathbf S_{<t}^l$ outside its top-$K_h$ principal support. Since
$\mathbf P_t^{l,R}$ lies in this complementary space, the accumulated
response of
$\Delta\mathbf W_t^{l,R}
=\mathbf B_t^{l,R}(\mathbf P_t^{l,R})^\top$
on historical layer inputs satisfies
\begin{equation}
    \sum_{k<t}
    \left\|
    \mathbf X_k^l
    \mathbf P_t^{l,R}
    (\mathbf B_t^{l,R})^\top
    \right\|_F^2
    \leq
    \rho_{K_h+1}^l
    \|\mathbf B_t^{l,R}\|_F^2.
    \label{eq:residual_historical_response}
\end{equation}
Thus, the residual mode simultaneously maximizes the remaining
current-task learning signal and bounds its response on historical
activations. \textcolor{DuLBEOrange}{\textbf{This explains why DuLBE can allocate a larger rank $r_R$ to
task-specific plasticity while keeping the historically sensitive shared
mode compact.}}

\paragraph{Gradient routing.}
Let
$\mathbf g_t^{l,R}
=\nabla_{\mathbf B_t^{l,R}}
\mathcal L_{\mathrm{c\text{-}ce}}$
denote the useful current-task gradient in the residual branch. The
bidirectional KL objective instead preserves the teacher's historical
semantic structure, and its systematic contribution is handled by the
shared branch that rewrites historically occupied directions. Under this
intended decomposition, the remaining mini-batch KL component in the
residual branch mainly reflects finite-sample variation, imperfect
subspace estimation, and nonlinear cross-layer coupling. We denote this
component by $\boldsymbol\xi_t^{l,R}$ and model it as
\begin{equation}
    \mathbb E[\boldsymbol\xi_t^{l,R}]=\mathbf0,
    \qquad
    \mathbb E[
    \|\boldsymbol\xi_t^{l,R}\|_F^2]
    =
    (\sigma_t^{l,R})^2.
    \label{eq:residual_kl_noise}
\end{equation}
If this component were routed to the residual branch, its update can
become
$\mathbf B_{t,+}^{l,R}
=\mathbf B_t^{l,R}
-\eta(\mathbf g_t^{l,R}
+\lambda\boldsymbol\xi_t^{l,R})$.

Assuming that $\mathcal L_{\mathrm{c\text{-}ce}}$ is locally
$L_l$-smooth with respect to $\mathbf B_t^{l,R}$, the expected
current-task loss after this noisy update satisfies
\begin{equation}
\begin{aligned}
    \mathbb E[
    \mathcal L_{\mathrm{c\text{-}ce}}
    (\mathbf B_{t,+}^{l,R})]
    \leq\;&
    \mathcal L_{\mathrm{c\text{-}ce}}
    (\mathbf B_t^{l,R})
    -
    \eta
    \left(
    1-\frac{L_l\eta}{2}
    \right)
    \|\mathbf g_t^{l,R}\|_F^2 \\
    &+
    \frac{L_l\eta^2\lambda^2}{2}
    (\sigma_t^{l,R})^2.
\end{aligned}
\label{eq:residual_noisy_descent}
\end{equation}
The final positive term is introduced solely by residual KL noise. It
weakens the guaranteed current-task loss decrease and perturbs the
dominant task-specific directions selected by
Eq.~\eqref{eq:optimal_residual_basis}. Consequently, we choose to route the KL
gradient only to $\mathbf B_t^{l,S}$, where historical-structure
preservation is required, while updating $\mathbf B_t^{l,R}$ only with
the current-task objective to preserve efficient plastic exploration.

\subsection{Analysis of Bridge-Prototype Ensemble}
\label{sec:bridge_analysis}

\begin{figure}[t]
    \centering
    \includegraphics[width=\linewidth]
    {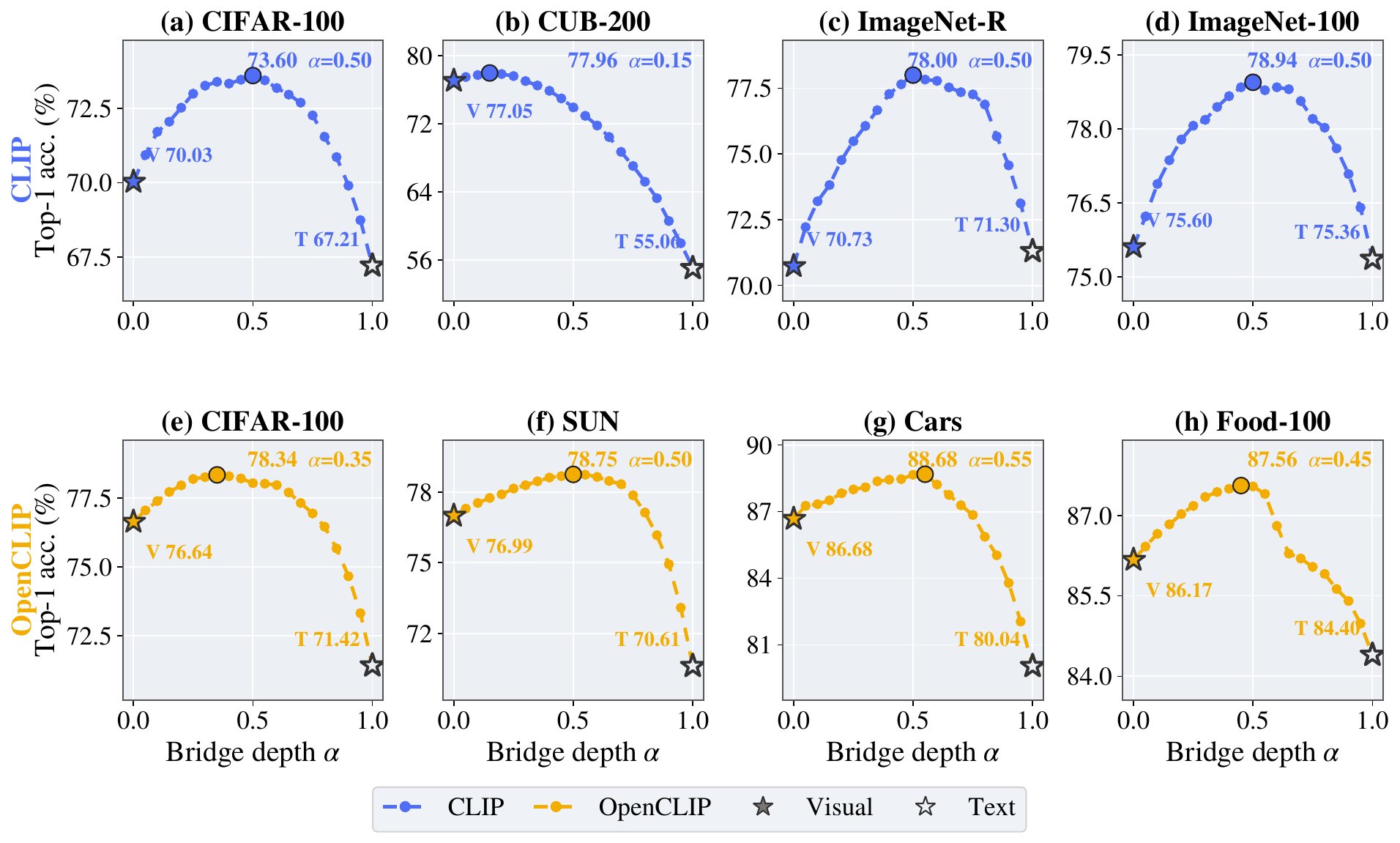}
    \caption{Dataset-level top-1 accuracy across bridge depths using
    frozen OpenAI CLIP (top) and OpenCLIP (bottom). Filled and hollow
    stars denote the visual and textual endpoints, respectively.}
    \label{fig:evidence_bridge_1}
\end{figure}

\begin{figure*}[t]
    \centering
    \includegraphics[width=\linewidth]
    {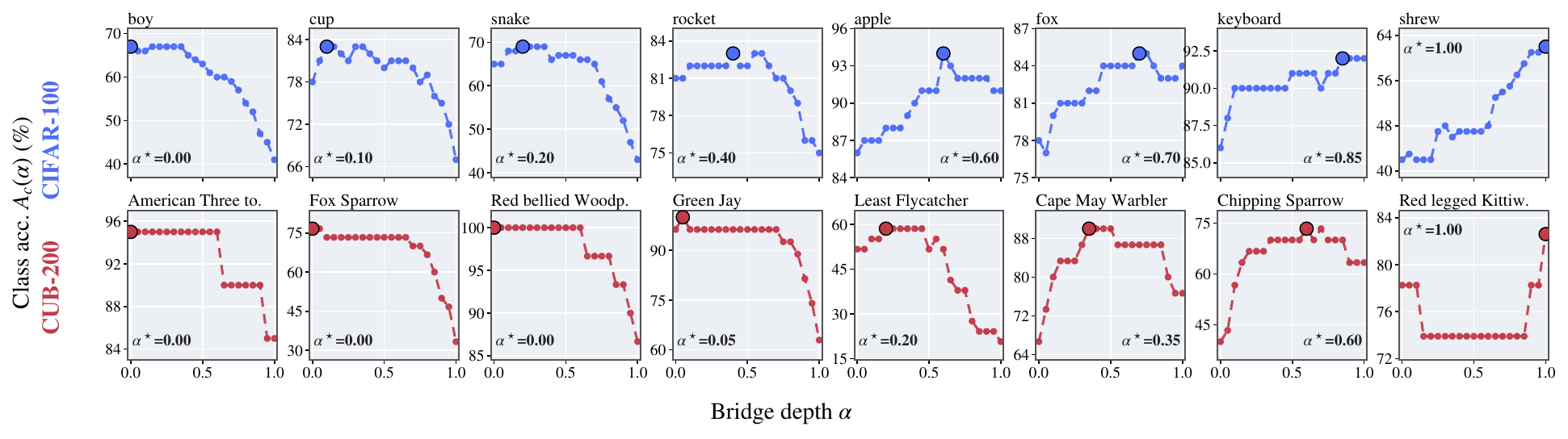}
    \caption{Class-wise accuracy across bridge depths for representative
    CIFAR (top) and CUB(bottom) classes. Each class is evaluated
    against all classes in the corresponding dataset. Enlarged markers
    indicate the class-specific optimal depths.}
    \label{fig:evidence_bridge_2}
\end{figure*}

\paragraph{Dataset and class-level evidence.}
The bridge-prototype ensemble is motivated by the fact that visual and
text prototypes provide complementary class information, while their
relative importance varies across scenarios and classes. We reveal this kind of effect using frozen CLIP and vary only the bridge depth
$\alpha$, where $\alpha=0$ and $\alpha=1$ denote the visual and textual
endpoints, respectively. As shown in
Figure~\ref{fig:evidence_bridge_1}, all 8 scenarios achieve their best
accuracy at an intermediate depth, with $\alpha^*$ ranging from $0.15$
to $0.55$. The best bridge exceeds the stronger endpoint by
$0.91$--$6.70$ percentage points, demonstrating that \textcolor{DuLBEOrange}{\textbf{the geodesic
interior contains useful discriminative representations beyond either
modality alone.}} Figure~\ref{fig:evidence_bridge_2} further shows that
classes within the same dataset prefer substantially different depths,
spanning visual-dominant, intermediate, and text-dominant regions. These
results motivate estimating bridge reliability separately for each class
instead of adopting a fixed global fusion depth.

\paragraph{Analysis of Geodesic Bridge Interpolation.}
The visual-text modality gap places visual and textual representations in
distinct regions of CLIP's shared embedding space
~\cite{liang2022mind,huang2025mind}. Consequently, a visual prototype and
its corresponding text embedding need not be geometrically aligned,
although they describe the same class. The visual prototype captures
appearance-specific evidence but may inherit specific-dependent bias,
whereas the text embedding provides stable semantics but may omit
fine-grained visual cues. Thus, the most discriminative class direction
need not coincide with either endpoint, but may lie between them, where
visual evidence and textual semantics are better integrated. The interior
accuracy gains in Figure~\ref{fig:evidence_bridge_1} and the
class-dependent optima in Figure~\ref{fig:evidence_bridge_2} provide
direct evidence for this non-endpoint discriminative region.

Geodesic interpolation provides a calibrated transition between these two
modality-specific semantic directions. Let
$\theta_c=\arccos(\mathbf v_c^\top\mathbf e_c)$. The resulting bridge
prototype satisfies
\begin{equation}
    \|\mathbf b_c(\alpha)\|_2=1,
    \qquad
    \angle(\mathbf v_c,\mathbf b_c(\alpha))
    =\alpha\theta_c.
\label{eq:bridge_calibrated_depth}
\end{equation}
Therefore, every bridge prototype remains on the CLIP unit hypersphere,
and $\alpha$ consistently represents the fraction of semantic
displacement from the visual modality toward the textual modality.
Prototypes of different classes at the same depth consequently retain
comparable value of logits.

For comparison, BOFA~\cite{li2025bofa} adopts ordinary linear
interpolation between the two modal endpoints:
\begin{equation}
\label{eq:linear_bridge_norm}
\begin{gathered}
\widetilde{\mathbf b}_c(\alpha)
=
(1-\alpha)\mathbf v_c+\alpha\mathbf e_c,\\[1mm]
\left\|\widetilde{\mathbf b}_c(\alpha)\right\|_2^2
=
1-2\alpha(1-\alpha)(1-\cos\theta_c).
\end{gathered}
\end{equation}
Its norm, and hence its logit scale, varies with the class-specific
modality gap $\theta_c$. Although normalization removes this scale
variation, the angular transition remains nonlinear, so the same
$\alpha$ represents different semantic progress across classes. In
contrast, geodesic interpolation preserves both unit norm and uniform
angular depth, making each depth a comparable all-class classifier.
Consequently, the reliability weights $\pi_c(\alpha_k)$ can identify the
most discriminative inter-modal region (depth) for each class semantics.

\section{Experimental Details}
\label{sec:experimental_setup}

\subsection{Datasets and Protocols}

\paragraph{Benchmark settings.}
We follow the two public protocols commonly used by the corresponding
baselines. \textbf{Setting-A} is the uniform configurations
\cite{huang2024rapf}\footnote{\url{https://github.com/linlany/RAPF}}
with OpenAI CLIP~\cite{radford2021learning} ViT-B/16. It contains
CIFAR-100~\cite{krizhevsky2009learning},
CUB-200-2011~\cite{wah2011caltech},
ImageNet-R~\cite{hendrycks2021many}, and 100-class subset of
ImageNet~\cite{deng2009imagenet}. We use the default public class orders.
ImageNet-R has no official train/test partition, we therefore use released random split, same as~\cite{huang2024rapf}.
\textbf{Setting-B} is the uniform 10-task configurations
\cite{zhou2025engine}\footnote{\url{https://github.com/LAMDA-CL/ICCV25-ENGINE}}
with OpenCLIP ViT-B/16~\cite{cherti2023reproducible} initialized from
\emph{laion400m\_e32} weights~\cite{schuhmann2021laion}. We choose
CIFAR-100, the 300-class SUN-397~\cite{xiao2010sun} subset, the 100-class
Stanford Cars~\cite{krause2013cars} subset, and the 100-class
Food-101~\cite{bossard2014food} subset. The class
order is shuffled with seed 1993, same as~\cite{zhou2025engine}.

\begin{table*}[t]
\centering
\caption{Dataset statistics and class-incremental protocols. ``Classes''
reports the number used by the protocol and the number in the original
benchmark. ``Stream'' gives the \textbf{number of tasks} $\times$
\textbf{new classes per task}, and ``Training template'' reports the text
prompt used during optimization. Train/test counts refer only to the
classes used in the CIL stream.}
\label{tab:dataset_protocols}
\small
\setlength{\tabcolsep}{6pt}
\renewcommand{\arraystretch}{1.12}
\begin{tabular}{@{}llrrrrl@{}}
\toprule
Setting & Dataset & Classes & Train & Test & Stream &
Training template \\
\midrule
A & CIFAR-100
  & 100/100 & 50,000 & 10,000 & 10 \texttimes{} 10
  & ``a good photo of a $c$.'' \\
A & CUB-200
  & 200/200 & 5,994 & 5,794 & 10 \texttimes{} 20
  & ``a good photo of a $c$.'' \\
A & ImageNet-R
  & 200/200 & 24,000 & 6,000 & 10 \texttimes{} 20
  & ``a good photo of a $c$.'' \\
A & ImageNet-100
  & 100/1,000 & 129,395 & 5,000 & 10 \texttimes{} 10
  & ``a good photo of a $c$.'' \\
\midrule
B & CIFAR-100
  & 100/100 & 50,000 & 10,000 & 10 \texttimes{} 10
  & ``a photo of a $c$.'' \\
B & SUN-397
  & 300/397 & 15,000 & 15,000 & 10 \texttimes{} 30
  & ``a photo of a $c$.'' \\
B & Stanford Cars
  & 100/196 & 4,114 & 4,062 & 10 \texttimes{} 10
  & ``a photo of a $c$.'' \\
B & Food-101
  & 100/101 & 75,000 & 25,000 & 10 \texttimes{} 10
  & ``a photo of $c$, a type of food.'' \\
\bottomrule
\end{tabular}
\end{table*}

\paragraph{Dataset.}
CIFAR-100 contains 32$\times$32 natural images from 100 object classes,
with 500 training and 100 test images per class. CUB-200 and Stanford
Cars are fine-grained benchmarks for 200 bird species and 196 car
make/model/year categories, respectively. ImageNet-R contains artistic
and non-photographic renditions of 200 ImageNet classes, whereas
ImageNet-100 retains natural images from 100 ImageNet classes. SUN-397
covers diverse indoor and outdoor scenes, with 50 training and 50 test
images per class in its standard partitions. Food-101 contains 750
training and 250 test images for each of 101 dishes. For Setting-B, only
the retained classes in Table~\ref{tab:dataset_protocols} enter the CIL
stream,and the remaining original classes are never used for training or
evaluation.

{
\paragraph{Text templates.}
Setting-A follows~\cite{huang2024rapf} and uses the single template
$\mathcal T_A=$ ``a good photo of a $c$.'' for every class name $c$.
Setting-B follows the actual configuration of \cite{zhou2025engine}, which selects
only the first entry of each dataset-specific template list. Thus,
CIFAR-100, SUN-397, and Stanford Cars use ``a photo of a $c$.'', while
Food-101 uses ``a photo of $c$, a type of food.'' The exact template for each benchmark is reported in
Table~\ref{tab:dataset_protocols}.
}

\paragraph{Image preprocessing.}
All images are mapped to 224$\times$224 and normalized by the preprocessing
pipeline associated with the corresponding pretrained CLIP model. We keep
the class subsets, class orders, train/test partitions, and preprocessing
fixed to the public configurations of \cite{huang2024rapf} and \cite{zhou2025engine}.

\paragraph{Data access in CIL.}
DuLBE uses only current-task data, without storing old
exemplars or introducing external data. We retain the original resource
settings of the compared methods. PROOF~\cite{zhou2025proof} stores 20
exemplars per observed class. CLG-CBM~\cite{yu2025language},
ENGINE~\cite{zhou2025engine}, and DesCLIP~\cite{he2026desclip} do not
replay old images but use externally generated language knowledge, such
as LLM-derived class concepts or attribute descriptions. In addition,
LfI~\cite{gong2026learning} uses
COCO Captions~\cite{chen2015microsoft} as an external
reference set during token generation, mixing 512 target samples with
512 COCO image-caption pairs per batch.
\subsection{Introduction to Compared Methods}
\label{sec:compared_methods}

We provide further descriptions of the compared methods in the main. Following the standard CLIP-based CIL evaluation protocol,
DualPrompt and CODA-Prompt are applied only to the visual branch of
CLIP, while the other methods are implemented according to their
original designs.

\begin{itemize}
    \setlength{\itemsep}{2pt}
    \setlength{\parsep}{0pt}
    \setlength{\parskip}{0pt}

    \item \textbf{ContinualCLIP~\cite{thengane2022clip}:}
    ContinualCLIP keeps CLIP frozen and performs zero-shot
    classification using the text embeddings of all classes observed so far. It requires neither continual optimization nor exemplar replay.

    \item \textbf{DualPrompt~\cite{wang2022dualprompt}:}
    DualPrompt learns complementary general and expert prompts for a frozen pretrained transformer. The general prompts encode
    task-shared knowledge, whereas the expert prompts capture
    task-specific information. 

    \item \textbf{CODA~\cite{smith2023coda}:}
    CODA decomposes prompts into learnable components and
    combines them using input-conditioned attention. This enables
    end-to-end prompt construction without replaying historical data.

    \item \textbf{PROOF~\cite{zhou2025proof}:}
    PROOF freezes the pretrained visual and text towers and expands
    task-specific projection layers as new classes arrive. Previously
    learned projections are fixed, while a cross-modal fusion module
    integrates projected visual features and visual-textual prototypes.

    \item \textbf{RAPF~\cite{huang2024rapf}:}
    RAPF adopts text semantics to estimate influence of new classes
    on old learned classes and adaptively adjusts their
    representations. It further employs decomposed parameter fusion to
    consolidate task-wise adapter updates and prevent forgetting.

    \item \textbf{SGCL~\cite{yu2024exploiting}:}
    SGCL exploits semantic relations encoded by the pretrained CLIP text
    tower. It combines semantically guided representation learning
    with semantically guided knowledge distillation to transfer class
    relations within and across tasks.

    \item \textbf{CLG-CBM~\cite{yu2025language}:}
    CLG-CBM introduces a language-guided concept bottleneck model (CBM) between
    frozen CLIP features and a linear classifier. Concept alignment improves
    interpretability, while semantic-guided prototype augmentation
    generates old-class pseudo-features to alleviate forgetting.

    \item \textbf{ENGINE~\cite{zhou2025engine}:}
    ENGINE injects external knowledge through visual and textual
    branches. The visual branch enriches features through data
    augmentation, while the textual branch introduces discriminative
    descriptions. The resulting knowledge is further exploited by
    re-ranking predictions using abundant text prompts at inference time.

    \item \textbf{DesCLIP~\cite{he2026desclip}:}
    DesCLIP constructs robust vision-general attribute-class associations using
    general attribute descriptions. An anchor-based filter selects
    vision-relevant descriptors, which guide visual adaptation and the
    calibration of class-text embeddings.

    \item \textbf{MG-CLIP~\cite{huang2025mind}:}
    MG-CLIP treats the intrinsic visual-textual modality gap as an
    indicator of retained pretrained cross-modal knowledge. It preserves the
    modality gap to reduce forgetting and compensates using visual learnable classifier in the visual space for CIL prediction.

    \item \textbf{BOFA~\cite{li2025bofa}:}
    BOFA adapts only CLIP's cross-modal bridge layer(projection from raw visual space to aligned visual space) and constrains its
    low-rank updates to a subspace approximately orthogonal to
    historical features. It further combines stable textual prototypes
    with adapted visual prototypes using linear Interpolation for classification.

    \item \textbf{LoDA-CLIP~\cite{he2026loda}:}
Adopts the official CLIP implementation of LoDA (Low-rank Decomposition and Adaptation) reported in
the original paper. LoDA constructs general and isolated LoRA down-projection
spaces from layer-wise feature statistics: the former captures
directions shared across previous and current tasks, while the latter
maximizes the current-to-historical feature energy ratio.
After optimizing the up-projections and recalibrating the general
update, the official CLIP implementation combines a frozen CLIP branch
with the adapted CLIP, fusing their prediction scores at inference.

    \item \textbf{LfI~\cite{gong2026learning}:}
    LfI mines the internal knowledge of CLIP without relying on an
    external captioning model. It constructs pseudo-captions by
    optimizing learnable tokens and performs adaptive mutual
    distillation between the textual classifier and a temporary visual
    classifier.
\end{itemize}

\subsection{Additional Implementation Details}

This subsection supplements the implementation settings reported in the
main paper with the dataset-wise training schedules and further details
of prospective-gradient estimation.

\paragraph{Training schedule and text adaptation.}
We train the CIL model on CIFAR for 2 epochs per task under both
OpenAI CLIP and OpenCLIP. CUB, ImageNet-100, and ImageNet-R are
trained for 4 epochs per task, whereas SUN, Cars, and Food under
OpenCLIP are trained for 10 epochs per task. All runs use Adam with a
cosine-annealing schedule, without warm-up, weight decay, or
mixed-precision training. Following~\cite{huang2025mind}, we adapt the
text tower using rank-8 LoRA while keeping its pretrained weights frozen.
The text-side LoRA parameters are optimized jointly with the visual
learner. Since the text tower serves as a stable semantic anchor, we do
not apply dual-mode allocation or historical-structure regularization to
its LoRA parameters.

\paragraph{Prospective-gradient estimation.}
At the beginning of each task, before initializing the new dual-mode
branches, we keep the model at its pre-task state and perform one
gradient-probing pass over the current-task training data. We use the
same cross-modal CE objective and current-class text embeddings as in
subsequent training. The mini-batch gradients are accumulated for all
adapted visual layers. No optimizer step is performed and no model
parameters are modified during this pass. The resulting gradients are
used only to construct the shared and residual bases and are then
discarded. For the first task, the bases are initialized from the
dominant gradient directions. For later tasks, the prospective gradients
are combined with the accumulated historical statistics for mode
allocation. This procedure uses only current-task samples and introduces
no historical replay.

Overall, our optimization epochs do not exceed those of the corresponding
standard protocols. Therefore, the performance gains of DuLBE do not
rely on an enlarged optimization budget.

\subsection{Evaluation Metrics}

After learning task $t$, the model is evaluated on all classes observed
so far, $\mathcal C_{\leq t}$, without task identity. We denote $A_t$ as
the resulting top-1 accuracy, which jointly reflects learning of new
classes and retention of previous ones. For the stream of $T$ tasks, we compute
$\bar{\mathcal A}=T^{-1}\sum_{t=1}^{T}A_t$. It averages accuracy across
all incremental stages and measures whether the model performs consistently throughout the learning process. In addition, we define $\mathcal A_l=A_T$, which evaluates the final model on all
benchmark classes after the complete task stream. It summarizes the
final balance between learning new classes and preserving old knowledge.

\section{Additional Experimental Results}
\label{sec:additional_results}

\subsection{Comparison with Other Baselines}
\label{sec:additional_baselines}

To complement the standard-protocol comparison in the main paper, \textcolor{DuLBEOrange}{\textbf{we
extend the evaluation along two dimensions: a) access to additional
training data and b) the choice of continual low-rank adaptation strategy.}}
All comparisons use Setting-A with the same OpenAI CLIP ViT-B/16
backbone, 10-task class splits, and evaluation metrics as in the main
paper.

\paragraph{Comparison with methods using additional training data.}
We compare DuLBE with methods that exploit additional training data
through either historical replay or external reference datasets.
SGCL~\cite{yu2024exploiting},
CLAP4CLIP~\cite{jha2024clap},
PROOF~\cite{zhou2025proof}, and
SPU$^{*}$~\cite{zhang2024overcoming}
retain 20 historical images per observed class. In contrast,
ZSCL~\cite{zheng2023preventing} distills knowledge from the 28k-image
validation set of Conceptual Captions~\cite{sharma2018conceptual},
whereas LfI~\cite{gong2026learning} samples 512 reference pairs from
COCO Captions~\cite{chen2015microsoft} in each token-generation batch.
We additionally report the replay-free SPU as a resource-matched
baseline. DuLBE uses neither historical exemplars nor external training
images and accesses only the current-task data.

\begin{table*}[t]
\centering
\small
\setlength{\tabcolsep}{6pt}
\renewcommand{\arraystretch}{1.1}
\begin{tabular*}{\textwidth}
{@{\extracolsep{\fill}}llcccccc@{}}
\toprule
\multicolumn{1}{c}{
    \raisebox{-2.2ex}[0pt][0pt]{Method}}
& \multicolumn{1}{c}{
    \raisebox{-2.2ex}[0pt][0pt]{Additional Data}}
& \multicolumn{6}{c}{Setting-A~(CLIP)} \\
\cmidrule(l){3-8}
& & \multicolumn{2}{c}{CIFAR}
& \multicolumn{2}{c}{CUB}
& \multicolumn{2}{c}{I.N.-R} \\
& & $\bar{\mathcal A}$ & $\mathcal A_l$
& $\bar{\mathcal A}$ & $\mathcal A_l$
& $\bar{\mathcal A}$ & $\mathcal A_l$ \\
\midrule
SGCL~\cite{yu2024exploiting}
& Replay (20/class)
& 89.1 & 82.7
& \textbf{87.1} & \textbf{82.9}
& 86.8 & 81.9 \\
CLAP4CLIP~\cite{jha2024clap}
& Replay (20/class)
& 85.1 & 76.4
& 85.2 & 79.9
& 85.0 & 79.2 \\
PROOF~\cite{zhou2025proof}
& Replay (20/class)
& 86.2 & 76.3
& -- & --
& 82.8 & 77.1 \\
SPU$^{*}$~\cite{zhang2024overcoming}
& Replay (20/class)
& \underline{89.2} & \underline{83.6}
& 81.0 & 70.3
& 85.7 & 80.1 \\
\midrule
ZSCL~\cite{zheng2023preventing}
& External (CC val., 28k)
& 82.4 & 74.1
& 71.1 & 60.6
& 85.5 & 79.7 \\
LfI~\cite{gong2026learning}
& External (COCO, 512/batch)
& 87.8 & 82.7
& -- & --
& \underline{88.1} & \underline{84.0} \\
SPU~\cite{zhang2024overcoming}
& None
& 84.9 & 76.4
& 75.9 & 68.2
& 83.1 & 77.5 \\
\midrule
\textbf{DuLBE}~(Ours)
& None
& \textbf{89.5} & \textbf{84.2}
& \underline{86.1} & \underline{80.6}
& \textbf{88.4} & \textbf{84.3} \\
\bottomrule
\end{tabular*}
\caption{Comparison with methods using additional training data under
Setting-A. Replay methods retain 20 historical images per observed
class. CC and COCO denote Conceptual Captions and COCO Captions,
respectively. SPU$^{*}$ is the replay-based variant, whereas SPU and
DuLBE use neither historical exemplars nor external training images.
The best results are highlighted in \textbf{bold}, and the second-best
results are \underline{underlined}.}
\label{tab:resource_baselines}
\end{table*}

Despite using no additional training data, DuLBE achieves the best
average and final accuracies on both CIFAR and ImageNet-R, while
ranking second on both metrics for CUB. It surpasses the strongest
replay-based results on CIFAR by $0.3$ and $0.6$ points,
respectively, and slightly outperforms LfI on ImageNet-R. On the fine-grained CUB benchmark, DuLBE remains
competitive with SGCL while avoiding the storage of old
exemplars.

\paragraph{Comparison with LoRA-based methods.}
We further compare DuLBE with representative LoRA-based continual learning methods: O-LoRA~\cite{wang2023olora},
InfLoRA~\cite{liang2024inflora},
CL-LoRA~\cite{he2025cllora}, and
LoDA~\cite{he2026loda}.
We follow the original rank settings and additionally evaluate LoDA
with $r=8$. For a fair comparison, all methods insert LoRA
modules into the same key and value projections and MLP layers of the
visual tower. Importantly, all reported LoRA baseline results use the
same standard CLIP text classifier for prediction, which is also adopted
by DuLBE$_{\mathrm{Text.}}$ for a classifier-matched comparison. These
baselines are formulated for generic pretrained models and are largely
agnostic to CLIP's dual-tower structure. In contrast, DuLBE
is specifically designed for CLIP-based continual learning
by coordinating visual adaptation with visual-textual knowledge
preservation. DuLBE$_{\mathrm{B.E.}}$ further reports the complete
method with the bridge-prototype ensemble classifier.

\begin{table}[!t]
\centering
\small
\setlength{\tabcolsep}{2pt}
\renewcommand{\arraystretch}{1.08}
\begin{tabularx}{\columnwidth}
{@{}>{\raggedright\arraybackslash}Xcccc@{}}
\toprule
\multirow{2}{*}{Method}
& \multicolumn{2}{c}{CIFAR}
& \multicolumn{2}{c}{I.N.-100} \\
\cmidrule(lr){2-3}\cmidrule(l){4-5}
& $\bar{\mathcal A}$ & $\mathcal A_l$
& $\bar{\mathcal A}$ & $\mathcal A_l$ \\
\midrule
O-LoRA ($r=8$/task)
& 86.2 & 77.3 & 83.4 & 72.0 \\

InfLoRA ($r=10$)
& 87.8 & 80.9 & 85.4 & 75.5 \\

CL-LoRA ($r=10$)
& 83.9 & 75.1 & 83.7 & 73.4 \\

LoDA ($r=8$)
& 86.0 & 80.4 & 85.6 & 74.7 \\
LoDA ($r=32$)
& 86.4 & 81.1 & 85.8 & 75.7 \\
\midrule
DuLBE$_{\mathrm{Text.}}$ ($r=1+8$)
& \underline{89.0} & \underline{83.0}
& \underline{87.5} & \underline{78.4} \\
\textbf{DuLBE}$_{\mathrm{B.E.}}$ ($r=1+8$)
& \textbf{89.5} & \textbf{84.2}
& \textbf{88.3} & \textbf{80.7} \\
\bottomrule
\end{tabularx}
\caption{Comparison with LoRA-based methods under Setting-A. All
methods use identical insertion locations in the visual tower, including
the key and value projections and the MLP layers. All LoRA baselines
and DuLBE$_{\mathrm{Text.}}$ use the same standard CLIP text classifier,
whereas DuLBE$_{\mathrm{B.E.}}$ uses the complete reliability-guided
bridge-prototype ensemble. For O-LoRA, $r=8$ denotes the rank of each
task-wise adapter. Best and second-best results are shown in
\textbf{bold} and \underline{underline}, respectively.}
\label{tab:lora_baselines}
\end{table}

\FloatBarrier

\subsection{Performance under the Learning-from-Half Protocol}
\label{sec:half_start}

We further evaluate DuLBE under the learning-from-half (LFH) protocol
in Setting-B. The initial stage contains half of the benchmark classes
and is followed by five equal incremental stages. Specifically, the
100-class benchmark splits of CIFAR-100, Stanford Cars, and Food-101
use B50 Inc10, while the 300-class SUN-397 split uses B150 Inc30.
All results are based on OpenCLIP ViT-B/16 pretrained on LAION-400M,
using the same class order and random seed as in the main Setting-B
experiments.

Baseline results are taken from the corresponding original papers or
matched evaluations under this configuration. PROOF follows its original setting with 20 replay
exemplars per observed class, while LfI uses COCO reference pairs during
training. As shown in
Table~\ref{tab:half_start_results}, we achieves the best average and
final accuracies on both CIFAR and SUN. Our DuLBE achieve these
results without data replay or external reference images, whereas PROOF
and LfI use replay exemplars and COCO reference pairs, respectively.
These results demonstrate that DuLBE effectively preserves the knowledge
acquired from a large initial task while remaining adaptable to subsequent
increments, extending its advantages beyond the uniform 10-task protocol.

\begin{table*}[t]
\centering
\small
\setlength{\tabcolsep}{5.5pt}
\renewcommand{\arraystretch}{1.1}
\begin{tabular*}{\textwidth}
{@{\extracolsep{\fill}}lcccccccc@{}}
\toprule
\multicolumn{1}{c}{
    \raisebox{-2.2ex}[0pt][0pt]{Method}}
& \multicolumn{8}{c}{
    Setting-B~(OpenCLIP)-LFH} \\
\cmidrule(l){2-9}
& \multicolumn{2}{c}{
    \shortstack{CIFAR\\[-1pt]{ B50 Inc10}}}
& \multicolumn{2}{c}{
    \shortstack{SUN\\[-1pt]{ B150 Inc30}}}
& \multicolumn{2}{c}{
    \shortstack{Cars\\[-1pt]{ B50 Inc10}}}
& \multicolumn{2}{c}{
    \shortstack{Food\\[-1pt]{ B50 Inc10}}} \\
& $\bar{\mathcal A}$ & $\mathcal A_l$
& $\bar{\mathcal A}$ & $\mathcal A_l$
& $\bar{\mathcal A}$ & $\mathcal A_l$
& $\bar{\mathcal A}$ & $\mathcal A_l$ \\
\midrule
ContinualCLIP~\cite{thengane2022clip}
& 76.5 & 71.4
& 75.0 & 72.1
& 78.3 & 76.4
& 84.8 & 81.9 \\

DualPrompt~\cite{wang2022dualprompt}
& 80.1 & 72.6
& 79.4 & 73.0
& 76.9 & 67.6
& 80.0 & 72.8 \\

CODA-Prompt~\cite{smith2023coda}
& 78.7 & 71.6
& 80.4 & 74.2
& 75.1 & 64.2
& 81.0 & 74.1 \\
\midrule
PROOF~\cite{zhou2025proof}
& 82.9 & 78.9
& 80.7 & 77.5
& 90.5 & 89.5
& 87.5 & 84.7 \\

RAPF~\cite{huang2024rapf}
& 82.2 & 77.9
& 78.0 & 73.1
& 75.9 & 63.2
& 85.5 & 81.2 \\

ENGINE~\cite{zhou2025engine}
& 83.2 & 79.5
& 81.6 & 78.5
& 91.6 & 90.0
& 86.9 & 83.9 \\

MG-CLIP~\cite{huang2025mind}
& 85.6 & 81.3
& 26.6 & 12.6
& 84.6 & 79.6
& 28.9 & 12.5 \\

BOFA~\cite{li2025bofa}
& 83.7 & 80.1
& \underline{82.3} & 79.3
& \underline{92.3} & 90.9
& 86.3 & 83.7 \\

LfI~\cite{gong2026learning}
& \underline{87.5} & \underline{85.6}
& 81.9 & \underline{79.8}
& \textbf{92.7} & \textbf{91.9}
& \textbf{91.9} & \textbf{89.9} \\
\midrule
\textbf{DuLBE}~(Ours)
& \textbf{88.5} & \textbf{85.8}
& \textbf{82.9} & \textbf{80.1}
& 92.1 & \underline{91.0}
& \underline{91.1} & \underline{88.2} \\
\bottomrule
\end{tabular*}
\caption{Comparison under the learning-from-half (LFH) protocol in
Setting-B. The best results are highlighted in \textbf{bold}, and the
second-best results are \underline{underlined}.}
\label{tab:half_start_results}
\end{table*}

\subsection{Performance with CLIP ViT-L/14 Backbone}
\label{sec:vit_l14}

To evaluate the scalability of DuLBE to a larger vision-language
backbone, we replace the OpenAI CLIP~\cite{radford2021learning}
ViT-B/16 used in Setting-A with ViT-L/14 and conduct experiments on
CIFAR and ImageNet-R. We retain the same uniform 10-task CIL,
with 10 and 20 new classes introduced per task, respectively. The CIL
protocol remains unchanged.

\begin{table}[!t]
\centering
\small
\setlength{\tabcolsep}{2pt}
\renewcommand{\arraystretch}{1.05}
\begin{tabularx}{\columnwidth}
{@{}>{\raggedright\arraybackslash}Xcccc@{}}
\toprule
\multirow{2}{*}{Method}
& \multicolumn{2}{c}{CIFAR}
& \multicolumn{2}{c}{I.N.-R} \\
\cmidrule(lr){2-3}\cmidrule(l){4-5}
& $\bar{\mathcal A}$ & $\mathcal A_l$
& $\bar{\mathcal A}$ & $\mathcal A_l$ \\
\midrule
PROOF~\cite{zhou2025proof}
& 89.9 & 83.6
& 91.3 & 87.3 \\

CLAP4CLIP~\cite{jha2024clap}
& 87.9 & 84.9
& 92.1 & 88.6 \\

SLCA~\cite{zhang2023slca}
& 90.1 & 84.6
& 90.0 & 86.8 \\

RAPF~\cite{huang2024rapf}
& 90.3 & 85.3
& 92.0 & 88.3 \\
\midrule
L2P++~\cite{wang2022l2p}
& 85.7 & 77.9
& 90.5 & 86.7 \\

DualPrompt~\cite{wang2022dualprompt}
& 86.6 & 79.1
& 90.7 & 87.1 \\

CODA~\cite{smith2023coda}
& 85.8 & 78.7
& 89.1 & 84.6 \\

ContinualCLIP~\cite{thengane2022clip}
& 80.5 & 73.5
& 87.0 & 83.1 \\

APER-Adapter~\cite{zhou2025aper}
& 80.2 & 72.0
& 89.2 & 85.4 \\

MOE4CL~\cite{yu2024moe4cl}
& 91.0 & 85.8
& 93.3 & 90.4 \\

CLAP4CLIP$^{*}$~\cite{jha2024clap}
& 74.4 & 71.6
& 91.1 & 87.6 \\

MagMax~\cite{marczak2024magmax}
& 90.2 & 86.1
& 93.2 & 89.6 \\

MG-CLIP~\cite{huang2025mind}
& 91.8 & 87.0
& 93.7 & 91.1 \\
\midrule
\textbf{DuLBE}$_{\mathrm{Text.}}$~(Ours)
& \underline{93.2} & \underline{89.1}
& \underline{93.9} & \underline{91.7} \\

\textbf{DuLBE}$_{\mathrm{B.E.}}$~(Ours)
& \textbf{93.4} & \textbf{89.4}
& \textbf{94.2} & \textbf{92.3} \\
\bottomrule
\end{tabularx}
\caption{Comparison using OpenAI CLIP ViT-L/14 under the uniform
Setting-A protocol. CIFAR-100 and ImageNet-R are divided into ten tasks
with 10 and 20 new classes per task, respectively. Baseline results
follow the matched ViT-L/14 evaluation reported by~\cite{huang2025mind}. CLAP4CLIP$^{*}$ denotes its replay-free
variant. Text. and B.E. denote the conventional text classifier and
bridge-prototype ensemble, respectively. The best results are
highlighted in \textbf{bold}, and the second-best results are
\underline{underlined}.}
\label{tab:vit_l14_results}
\end{table}

These baseline results are all taken from the matched ViT-L/14 evaluation
reported by~\cite{huang2025mind}. PROOF and CLAP4CLIP use
real-data replay, whereas SLCA and RAPF use feature replay.
CLAP4CLIP$^{*}$ denotes the exemplar-free variant of CLAP4CLIP. In
contrast, DuLBE uses neither replay nor external reference
data. This experiment examines whether DuLBE's dual-mode adaptation
and bridge-prototype classification remain effective as the capacity
of the CLIP backbone increases. As shown in Table~\ref{tab:vit_l14_results}, DuLBE achieves the best
results across all four metrics, while its text-classifier variant
consistently ranks second. These results confirm that the dual-mode
learner remains effective with a larger CLIP backbone and that the
bridge-prototype ensemble provides further improvements.

\FloatBarrier

\subsection{Additional Ablation Studies}
\label{sec:additional_ablations}

We further analyze the bridge-prototype hyperparameters and the
insertion locations of the dual-mode low-rank learner.

\paragraph{Bridge-depth discretization and reliability smoothing.}
The number of bridge depths $K_b$ controls the discretization
granularity of the visual-text geodesic. $\beta$ controls the
smoothness of the class-wise reliability weights. We vary both
hyperparameters while keeping the trained dual-mode learner and all
other settings unchanged.

\begin{figure}[t]
    \centering
    \includegraphics[width=0.92\linewidth]
    {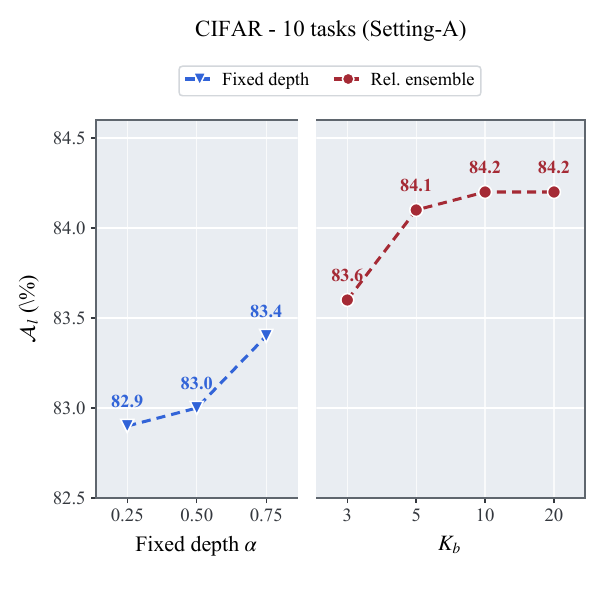}
    \caption{Effect of fixed bridge depth $\alpha$ and the number of
    reliability-weighted bridge depths $K_b$ on CIFAR-100 under
    Setting-A. Blue denotes a fixed-depth classifier, while red denotes
    the reliability-guided bridge-prototype ensemble.}
    \label{fig:bridge_depth_ablation}
\end{figure}

\begin{figure*}[t]
    \centering
    \includegraphics[width=0.8\textwidth]
    {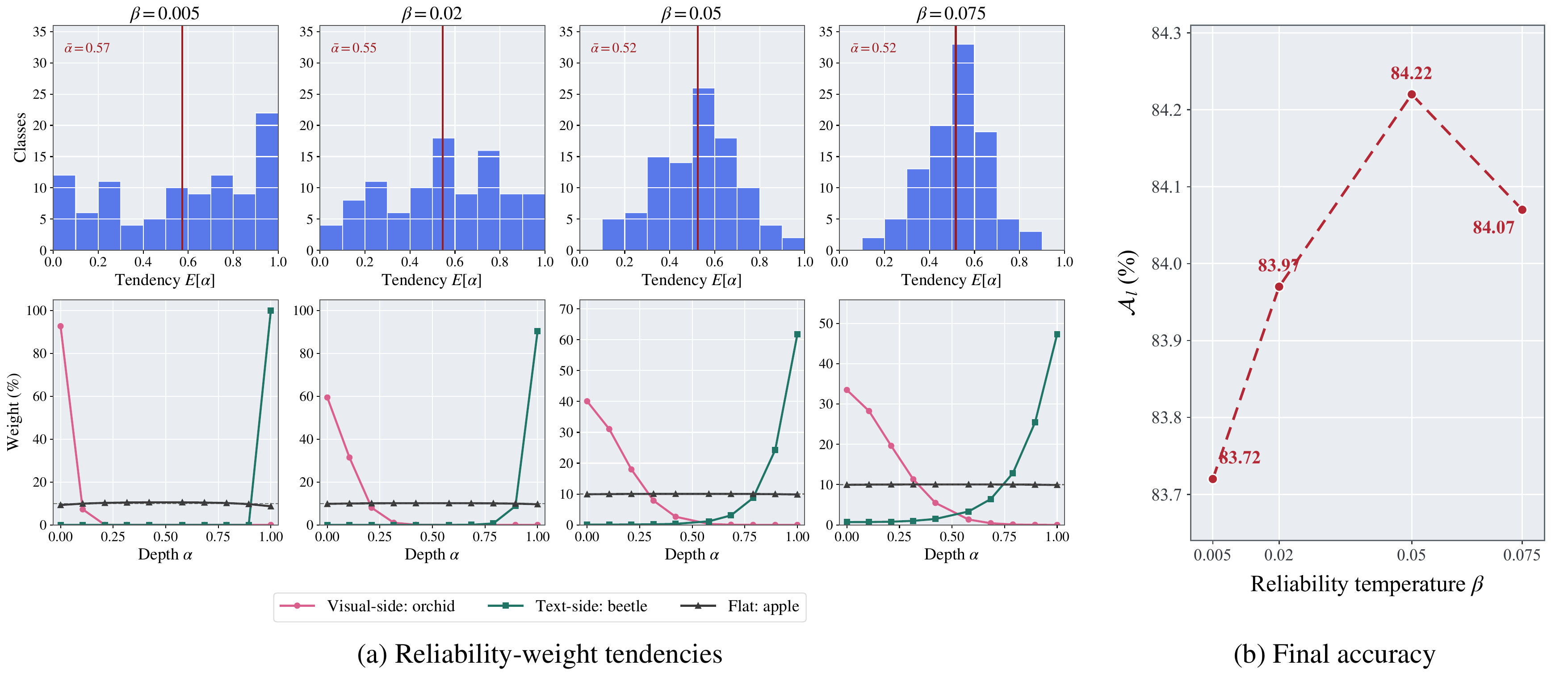}
    \caption{Effect of the reliability temperature $\beta$ with
    $K_b=10$ on CIFAR-100 under Setting-A.
    (a) Distributions of the class-wise expected bridge depth
    $\mathbb E_{\pi_c}[\alpha]$ (top) and reliability weights of three
    representative classes (bottom). The red vertical line denotes the
    mean expected depth across classes.
    (b) Final accuracy under different values of $\beta$.}
    \label{fig:bridge_beta_analysis}
\end{figure*}

As shown in Figure~\ref{fig:bridge_depth_ablation}, reliability-guided
ensembling consistently outperforms the fixed-depth classifiers.
Increasing $K_b$ from $3$ to $10$ improves $\mathcal A_l$ from
83.6\% to 84.2\%, whereas $K_b=20$ provides no further gain.
Thus, $K_b=10$ provides sufficient bridge granularity without
introducing redundant prototypes. Figure~\ref{fig:bridge_beta_analysis}(a) shows that a very small
$\beta$ produces nearly one-hot depth selection, making the ensemble
sensitive to reliability noise. Increasing $\beta$ smoothly distributes
weight over neighboring depths while preserving class-specific
preferences: \emph{orchid} favors the visual side, \emph{beetle}
favors the textual side, and \emph{apple} remains approximately
depth-insensitive. As shown in Figure~\ref{fig:bridge_beta_analysis}(b),
$\mathcal A_l$ increases from 83.72\% at $\beta=0.005$ to
84.22\% at $\beta=0.05$, before slightly decreasing to 84.07\%
at $\beta=0.075$. We therefore use $\beta=0.05$, which provides the
best balance between depth selectivity and reliability smoothing.

\paragraph{Insertion locations of the dual-mode low-rank learner.}
\begin{table}[!b]
\centering
\small
\setlength{\tabcolsep}{6pt}
\renewcommand{\arraystretch}{1.08}
\begin{tabular*}{\columnwidth}
{@{\extracolsep{\fill}}lcc@{}}
\toprule
Adapted visual modules
& $\bar{\mathcal A}\uparrow$
& $\mathcal A_l\uparrow$ \\
\midrule
$K,V$
& 88.8 & 83.0 \\
$Q,K,V$
& 89.1 & 83.9 \\
$K,V+\mathrm{MLP}$ (default)
& \underline{89.5} & \textbf{84.2} \\
$\mathrm{MLP}$
& 89.2 & 83.5 \\
$Q,K,V+\mathrm{MLP}$
& \textbf{89.6} & \underline{84.1} \\
\bottomrule
\end{tabular*}
\caption{Effect of the insertion locations of the dual-mode low-rank
learner on CIFAR-100 under Setting-A. Here, $Q$, $K$, and $V$ denote
the query, key, and value projections in visual self-attention, while
$\mathrm{MLP}$ denotes the feed-forward block. The same per-module ranks
are used for all configurations. The best results are highlighted in
\textbf{bold}, and the second-best results are
\underline{underlined}.}
\label{tab:insertion_location_ablation}
\end{table}

We compare different insertion locations within each visual Transformer
block, including the query ($Q$), key ($K$), and value ($V$)
projections and the feed-forward network ($\mathrm{MLP}$). For example,
$K,V+\mathrm{MLP}$ applies the dual-mode learner to the key and value
projections and the MLP layers. All configurations use
$r_S=1$ and $r_R=8$, while modules outside the specified locations
remain frozen.

As shown in Table~\ref{tab:insertion_location_ablation}, MLP-only
adaptation performs better than adapting only the key and value
projections, while combining them provides complementary benefits.
Including the query projection brings little additional improvement and
slightly weakens final retention. We therefore adopt
$K,V+\mathrm{MLP}$ as the default configuration, providing a concise
and unified adaptation scheme for both attention and feature
transformation.

\FloatBarrier

\subsection{Overhead}
\label{sec:overhead}

\paragraph{Training and inference overhead.}
Table~\ref{tab:overhead_cifar_a} reports the peak number of parameters
optimized within one task, including both the visual and text towers
and any learnable auxiliary classifier. Inference GFLOPs are measured
per image, following the convention that one ViT-B/16 visual forward
costs 17.60 GFLOPs. Class-text embeddings are computed once and cached
rather than repeatedly encoded for each test image. Mergeable LoRA
updates are folded into the pretrained weights before inference.
Additional parameters denote method-specific model parameters and
persistent prototype buffers retained beyond one standard CLIP model.
The common CLIP text-embedding cache is excluded.

\begin{table*}[!t]
\centering
\small
\setlength{\tabcolsep}{7pt}
\renewcommand{\arraystretch}{1.12}
\begin{tabular*}{\textwidth}{@{\extracolsep{\fill}}lccccc@{}}
\toprule
\multicolumn{1}{c}{\raisebox{-1.6ex}[0pt][0pt]{Method}}
& \multicolumn{1}{c}{Training Stage}
& \multicolumn{2}{c}{Inference Stage}
& \multicolumn{2}{c}{CIFAR (A)} \\
\cmidrule(lr){2-2}
\cmidrule(lr){3-4}
\cmidrule(l){5-6}
& \shortstack{Trainable\\Params (M)}
& \shortstack{GFLOPs\\(per sample)}
& \shortstack{Additional\\Params (M)}
& $\bar{\mathcal A}\uparrow$
& $\mathcal A_l\uparrow$ \\
\midrule
RAPF~\cite{huang2024rapf}
& 0.26 & 17.6003 & 0.262
& 86.2 & 79.0 \\

LoRA-CLIP$_{r=32}$~\cite{hu2022lora}
& 6.88 & 17.6001 & 0.000
& 86.2 & 79.1 \\

LoDA-CLIP$_{r=32}^{\dagger}$~\cite{he2026loda}
& 1.77 & 35.2001 & 86.244
& 86.4 & 81.1 \\

MG-CLIP~\cite{huang2025mind}
& 0.54 & 17.6001 & 0.051
& 87.0 & 80.6 \\

SGCL$^{*}$~\cite{yu2024exploiting}
& 7.09 & 17.6001 & 0.000
& 86.4 & 79.5 \\

LfI~\cite{gong2026learning}
& 149.62 & 17.6001 & 0.000
& 87.8 & \underline{82.7} \\
\midrule
\textbf{DuLBE}$_{r=1}$~(Ours)
& 0.75 & 17.6005 & 0.052
& \underline{88.5} & 82.6 \\

\textbf{DuLBE}$_{r=1+8}$~(Ours)
& 1.27 & 17.6005 & 0.052
& \textbf{89.5} & \textbf{84.2} \\
\bottomrule
\end{tabular*}

\caption{Training and inference overhead on CIFAR-100 under Setting-A.
Trainable parameters cover the complete method, including trainable
components in both CLIP towers. GFLOPs include visual encoding and
classification over all 100 classes. SGCL$^{*}$ denotes the replay-free variant. MG-CLIP uses one image encoding
and combines a text classifier with a learned visual classifier.
DuLBE directly evaluates $K_b=10$ bridge depths, incurring
$K_b|\mathcal C|d=0.000512$ GFLOPs for bridge scoring. Additional
parameters include method-specific model parameters and persistent
prototype buffers retained beyond the standard CLIP model, while
excluding the common text-embedding cache. The best results are in
\textbf{bold}, and the second-best results are
\underline{underlined}.}
\label{tab:overhead_cifar_a}
\end{table*}

\noindent
For LoDA-CLIP~\cite{he2026loda}, we follow its official CLIP extension,
which combines predictions from a frozen pretrained visual encoder and
a separately adapted visual encoder. Its additional parameters
therefore include one complete ViT-B/16 visual tower and the stored
visual prototypes. This differs from the single-backbone LoDA setting,
in which the low-rank updates can be merged without retaining
additional inference parameters.

\noindent
For DuLBE, the trainable-parameter count includes rank-8 text-side LoRA
modules applied to the key/value projections and MLP layers. The visual
trainable parameters are 0.0645M and 0.5806M for $r=1$ and $r=1+8$,
respectively. At inference, DuLBE retains only the class visual
prototypes and reliability weights, requiring
$|\mathcal C|d+|\mathcal C|K_b=0.0522$M additional parameters. The
bridge prototypes are generated from these parameters and the cached
text embeddings when needed.
\paragraph{Memory overhead.}
We analyze two complementary sources of persistent memory:
class-level representations and layer-wise historical statistics. Only
numerical states retained across tasks are counted. Pretrained CLIP
parameters, temporary tensors, and the class-text embedding cache shared
by CLIP-based classifiers are excluded.

\noindent{\textcolor{DuLBEOrange}{\textbf{Representation storage.}}}
RAPF~\cite{huang2024rapf} retains a feature mean and a full covariance
matrix for each historical class, while CLG-CBM
~\cite{yu2025language} additionally stores its selected concept
representations. Such detailed class distributions support feature-level
knowledge preservation, but their storage increases rapidly with the
number of classes. ENGINE~\cite{zhou2025engine} is more compact because
it combines class prototypes with shared GDA statistics. In contrast,
DuLBE stores only one visual prototype and a small set of reliability
weights per class. Its bridge prototypes are generated when needed and
therefore require no additional prototype bank. As shown in
Table~\ref{tab:representation_memory}, DuLBE requires only 0.20 MiB of
representation memory while achieving the highest final accuracy under
Setting-A.

\noindent\textcolor{DuLBEOrange}{\textbf{Layer-statistic storage.}}
The layer-wise statistics of DuLBE are retained only at the locations
where the dual-mode low-rank learner is inserted. Each statistic is
updated in place, so its storage is independent of the number of tasks
and observed classes. We evaluate the full configuration and three
compact variants under Setting-B using OpenCLIP ViT-B/16 pretrained on
LAION-400M, reporting final accuracy on CIFAR and Food.

Variant A applies the learner only to the key and value projections of
all visual blocks, reducing the statistic storage from 486.00 to
27.00 MiB while achieving final accuracies of 83.9\% and 85.5\%.
Variant B further restricts the learner to the last four visual blocks
and retains comparable performance with only 9.00 MiB. Variant C adapts
only the cross-modal bridge layer and uses the same 2.25 MiB statistic
budget as BOFA~\cite{li2025bofa}. Under this matched configuration,
DuLBE-C outperforms BOFA by 1.9 and 1.4 points on CIFAR-100 and
Food-101, respectively. \textcolor{DuLBEOrange}{\textbf{Overall, the three variants reduce statistic
storage by approximately 18$\times$, 54$\times$, and 216$\times$ while
retaining strong final performance, demonstrating that DuLBE can be
deployed flexibly under different memory budgets.}}

\FloatBarrier

\begin{table}[H]
\centering
\small
\renewcommand{\arraystretch}{1.08}
\begin{tabularx}{\columnwidth}
{@{}l>{\raggedright\arraybackslash}Xcc@{}}
\toprule
Method
& Stored state
& MiB$\downarrow$
& $\mathcal A_l\uparrow$ \\
\midrule
RAPF
& Mean + covariance
& 100.20
& 79.0 \\

ENGINE
& Prototype + GDA
& 3.03
& 73.1 \\

CLG-CBM
& Mean + covariance + concepts
& 102.15
& 76.9 \\

\textbf{DuLBE}
& Prototype + reliability
& \textbf{0.20}
& \textbf{84.2} \\
\bottomrule
\end{tabularx}
\caption{Representation memory and final accuracy on CIFAR-100 under
Setting-A.}
\label{tab:representation_memory}
\end{table}

\begin{table}[H]
\centering
\small
\renewcommand{\arraystretch}{1.08}
\begin{tabularx}{\columnwidth}
{@{}l>{\raggedright\arraybackslash}Xccc@{}}
\toprule
& & & \multicolumn{2}{c}{$\mathcal A_l\uparrow$} \\
\cmidrule(l){4-5}
Config.
& Learner location
& MiB$\downarrow$
& CIFAR
& Food \\
\midrule
Full
& $K,V+\mathrm{MLP}$
& 486.00
& \textbf{84.5}
& \textbf{86.7} \\

A
& $K,V$, all blocks
& 27.00
& \underline{83.9}
& \underline{85.5} \\

B
& $K,V$, last four
& 9.00
& 83.6
& 85.1 \\

C
& Bridge layer
& 2.25
& 81.2
& 84.1 \\
\midrule
BOFA
& Bridge layer
& 2.25
& 79.3
& 82.7 \\
\bottomrule
\end{tabularx}
\caption{Statistic storage and final accuracy under Setting-B.}
\label{tab:statistic_memory}
\end{table}

\begin{figure*}[!t]
    \centering
    \includegraphics[width=\linewidth]
    {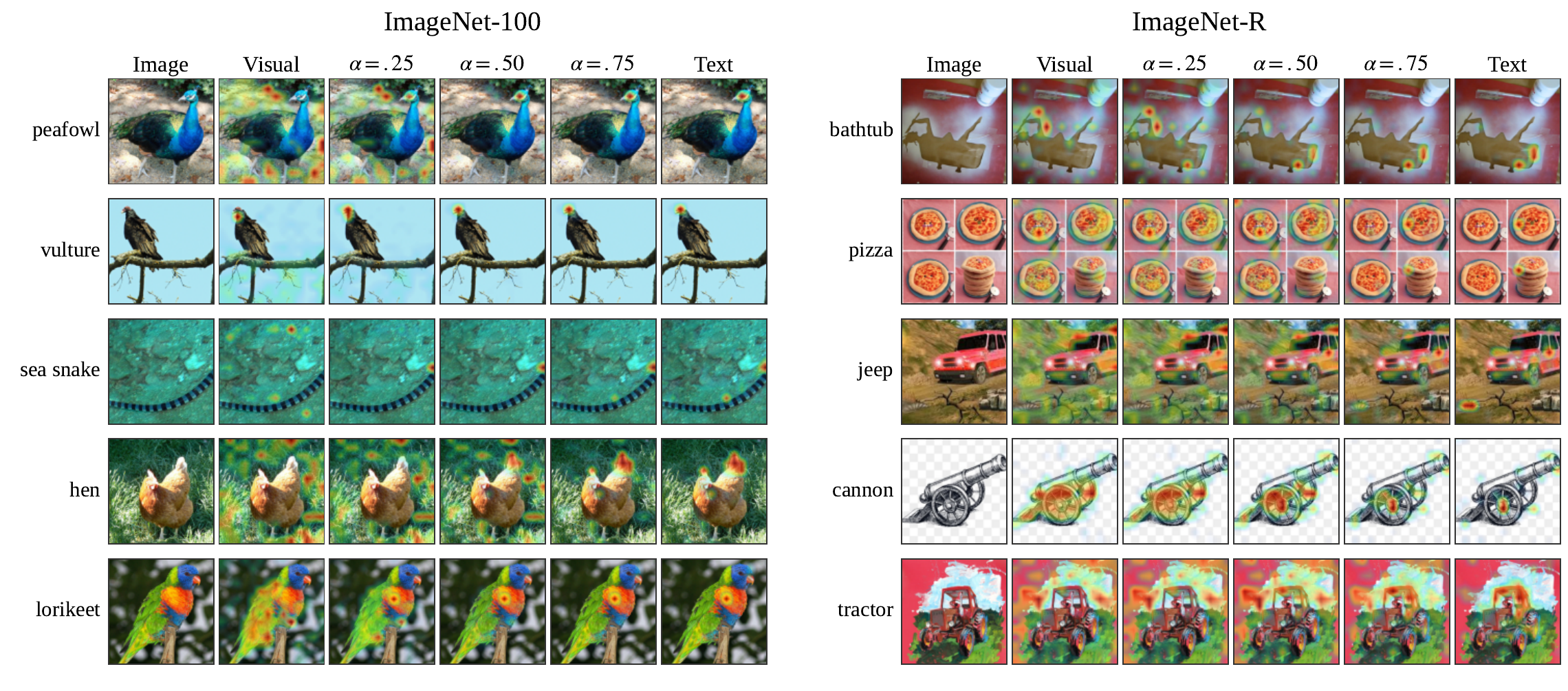}
    \caption{Prototype-conditioned Layer Grad$\times$Act maps on
    ImageNet-100 and ImageNet-R. Columns show the visual prototype,
    bridge prototypes at $\alpha\in\{0.25,0.50,0.75\}$, and the text
    embedding. Warmer colors indicate larger positive contributions to
    the image-prototype similarity.}
    \label{fig:attribution}
\end{figure*}

\begin{figure}[t]
    \centering
    \includegraphics[width=\linewidth]
    {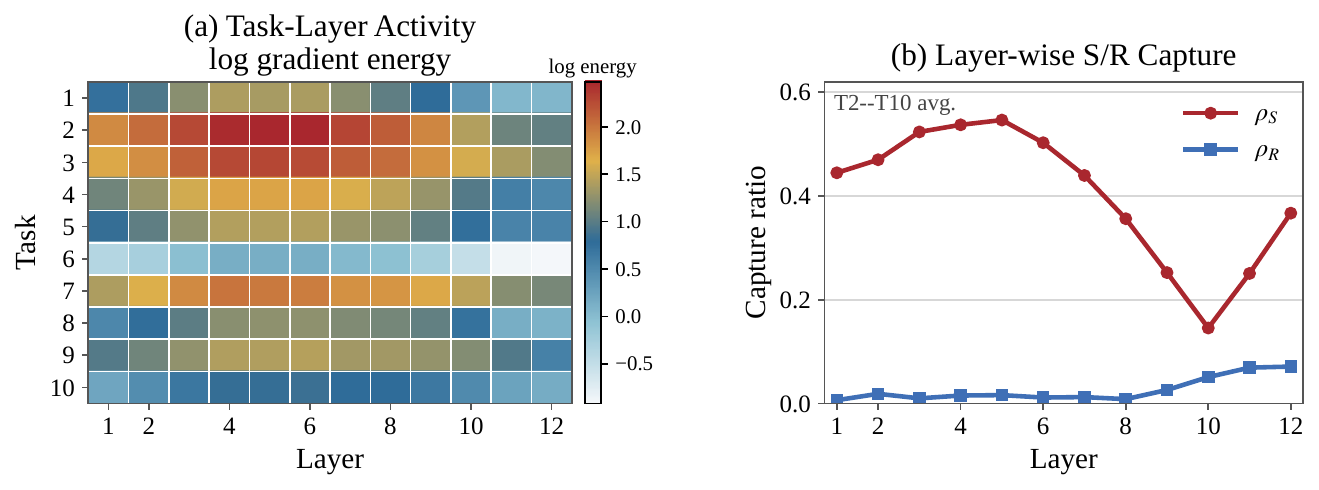}
    \caption{CIFAR-100 gradient-space diagnostics. (a) Task-layer
    log expected-gradient energy across visual Transformer layers.
    (b) Layer-wise expected-gradient capture ratios of the shared and
    residual spaces, averaged over Task2--Task10 (T2--T10). The two ratios are computed
    independently and are not complementary.}
    \label{fig:gradient_space}
\end{figure}

\subsection{Visualizations}
\label{sec:vis}

\paragraph{Bridge-depth attribution.}
We use prototype-conditioned Layer Grad$\times$Act
maps~\cite{zhao2024gradeclip} to trace the visual evidence associated
with each bridge depth. As shown in Figure~\ref{fig:attribution},
changing the conditioning prototype alters not only the attribution
strength but also its spatial distribution. The visual endpoint often
captures broader appearance and contextual evidence, whereas the text
endpoint tends to emphasize more localized, class-discriminative
regions. Intermediate bridge prototypes provide distinct transitions
between these patterns, although the transition varies across classes.
Similar behavior is observed on both ImageNet-100 and the
rendition-style ImageNet-R, suggesting that the intermediate prototypes
remain meaningful under visual domain shifts. These observations show
that different bridge depths provide complementary decision evidence,
supporting class-dependent reliability weighting rather than reliance
on a single fixed depth.

\paragraph{Gradient-space alignment.}
Following the gradient-demand allocation defined in the main paper,
Figure~\ref{fig:gradient_space} analyzes the prospective cross-modal
gradient $\mathbf G_t^l$ used to construct the two bases. Panel (a)
shows that its energy varies substantially across tasks and visual
layers, supporting the layer-wise allocation of low-rank directions
rather than using a fixed subspace for the entire visual tower.

For $t\geq2$, we also measure the gradient energy captured by the shared and
residual bases as
$\rho_S^l=\|\mathbf G_t^l\mathbf P_t^{l,S}\|_F^2/
\|\mathbf G_t^l\|_F^2$ and
$\rho_R^l=\|\mathbf G_t^l\mathbf P_t^{l,R}\|_F^2/
\|\mathbf G_t^l\|_F^2$, respectively. Panel (b) shows that the compact
shared basis $\mathbf P_t^{l,S}$ captures substantial demand in the
early and middle layers, confirming that a single direction selected
from the historical support $\mathcal H_{<t}^l$ can reuse transferable
visual structure. In deeper layers, its capture ratio decreases while
the residual ratio increases, indicating stronger demand outside the
historical support. This demand is captured by
$\mathbf P_t^{l,R}\subseteq \operatorname{span}((\mathcal H_{<t}^l)^\perp)$ to provide
low-interference task-specific plasticity. Since the two ratios measure
the energy captured by two compact bases of different ranks, they are
computed independently and do not form a complete decomposition of the
gradient.

\FloatBarrier

\section{Complete Procedure}
\label{sec:complete_procedure}

Algorithm~\ref{alg:dulbe} summarizes the complete task-wise procedure
of DuLBE.

\begin{algorithm2e*}[!t]
\caption{Procedure of \textsc{DuLBE}}
\label{alg:dulbe}
\small
\SetAlgoVlined
\SetAlgoNlRelativeSize{-1}
\SetKwInput{KwIn}{Input}
\SetKwInput{KwOut}{Output}
\SetKwFor{ForEach}{for each}{do}{end for}
\setlength{\FrameSep}{4pt}
\setlength{\OuterFrameSep}{3pt}

\KwIn{CLIP $\theta_0$; task stream
$\{\mathcal D_t,\mathcal C_t\}_{t=1}^{N}$; adapted layers $\mathcal V$;
hyperparameters $\Omega$.}
\KwOut{$\theta_N$; stored
$\{\mathbf v_c,\pi_c(\alpha_k)\}_{c,k}$.}

\textbf{Initialization:}
$\bar\theta_0\leftarrow\theta_0$ and
$\mathbf S_{<1}^l\leftarrow\mathbf0$,
$\forall l\in\mathcal V$\;

\For{task $t=1,\ldots,N$}{
    \colorlet{shadecolor}{DuLBEGray}
    \begin{shaded}
    \DuLBEModuleTitle{Gradient-Demand Dual-Mode Allocation}

    Estimate prospective demand:
    $\mathbf G_t^l\leftarrow
    \nabla_{\mathbf W_{t-1}^l}
    \mathcal L_{\mathrm{c\text{-}ce}}
    (\theta_{t-1};\mathcal D_t)$,
    $\forall l\in\mathcal V$\;

    \ForEach{adapted layer $l\in\mathcal V$}{
        Allocate modes:
        $(\mathbf P_t^{l,S},\mathbf P_t^{l,R})
        \leftarrow
        \operatorname{ModeAlloc}
        (\mathbf G_t^l,\mathbf S_{<t}^l)$\;

        Freeze $\mathbf P_t^{l,S},\mathbf P_t^{l,R}$ and initialize
        $\mathbf B_t^{l,S},\mathbf B_t^{l,R}
        \leftarrow\mathbf0$\;
    }
    \end{shaded}

    \colorlet{shadecolor}{DuLBEBlue}
    \begin{shaded}
    \DuLBEModuleTitle{Routed Cross-Modal Training}

    \ForEach{mini-batch $\mathcal B\subset\mathcal D_t$}{
        $(\mathbf g_{\mathrm{ce}}^{l,S},
        \mathbf g_{\mathrm{ce}}^{l,R})
        \leftarrow
        \nabla_{(\mathbf B_t^{l,S},\mathbf B_t^{l,R})}
        \mathcal L_{\mathrm{c\text{-}ce}}$,\quad
        $\mathbf g_{\mathrm{kl}}^{l,S}
        \leftarrow
        \mathbb I[t>1]
        \nabla_{\mathbf B_t^{l,S}}
        \mathcal L_{\mathrm{bi\text{-}kl}}
        (\bar\theta_{t-1})$\;

        Route and update:
        $\mathbf B_t^{l,S}
        \leftarrow
        \operatorname{OptStep}
        (\mathbf g_{\mathrm{ce}}^{l,S}
        +\lambda\mathbf g_{\mathrm{kl}}^{l,S})$,\quad
        $\mathbf B_t^{l,R}
        \leftarrow
        \operatorname{OptStep}
        (\mathbf g_{\mathrm{ce}}^{l,R})$\;
    }
    \end{shaded}

    \colorlet{shadecolor}{DuLBEYellow}
    \begin{shaded}
    \DuLBEModuleTitle{Reliability-Guided Bridge-Prototype Ensemble}

    \ForEach{new class $c\in\mathcal C_t$}{
        $\mathbf v_c\leftarrow
        \operatorname{Normalize}
        \left(\sum_{x_i\in\mathcal X_c^t}\mathbf z_i\right)$,\quad
        $\mathbf b_c(\alpha_k)\leftarrow
        \operatorname{Slerp}
        (\mathbf v_c,\mathbf e_c;\alpha_k)$\;

        $p_{\alpha_k}(j\mid x_i)\propto
        \exp\!\left(
        \tau\mathbf z_i^\top\mathbf b_j(\alpha_k)
        \right)$,\quad
        $\rho_c(\alpha_k)\leftarrow
        |\mathcal X_c^t|^{-1}
        \sum_{x_i\in\mathcal X_c^t}
        p_{\alpha_k}(c\mid x_i)$\;

        $\pi_c(\alpha_k)\leftarrow
        \operatorname{softmax}_{k}
        \bigl(\rho_c(\alpha_k)/\beta\bigr)$;
        store
        $\{\mathbf v_c,\pi_c(\alpha_k)\}_{k=1}^{K_b}$\;
    }

    Update historical state:
    $\mathbf S_{<t+1}^l\leftarrow
    \mathbf S_{<t}^l+
    (\mathbf X_t^l)^\top\mathbf X_t^l$,
    $\forall l\in\mathcal V$\;

    $\bar\theta_t\leftarrow
    \operatorname{StopGrad}(\theta_t)$;
    discard $\mathcal D_t$\;

    Evaluate after task $t$:
    $s_c(x)\leftarrow
    \sum_{k=1}^{K_b}
    \pi_c(\alpha_k)
    \tau\mathbf z(x)^\top\mathbf b_c(\alpha_k)$,\quad
    $\hat y(x)\leftarrow
    \arg\max_{c\in\mathcal C_{\leq t}}s_c(x)$\;
    \end{shaded}
}
\end{algorithm2e*}

\newpage

\begingroup
\small
\setlength{\bibsep}{0pt}
\ifdefined\DuLBEAppendixEmbedded

\else
\bibliography{references}

\begin{thebibliography}{40}
\providecommand{\natexlab}[1]{#1}

\bibitem[{Bossard, Guillaumin, and Van~Gool(2014)}]{bossard2014food}
Bossard, L.; Guillaumin, M.; and Van~Gool, L. 2014.
\newblock Food-101: Mining Discriminative Components with Random Forests.
\newblock In \emph{European Conference on Computer Vision}, 446--461.

\bibitem[{Chen et~al.(2015)Chen, Fang, Lin, Vedantam, Gupta, Doll{\'a}r, and
  Zitnick}]{chen2015microsoft}
Chen, X.; Fang, H.; Lin, T.-Y.; Vedantam, R.; Gupta, S.; Doll{\'a}r, P.; and
  Zitnick, C.~L. 2015.
\newblock Microsoft {COCO} Captions: Data Collection and Evaluation Server.
\newblock \emph{arXiv preprint arXiv:1504.00325}.

\bibitem[{Cherti et~al.(2023)Cherti, Beaumont, Wightman, Wortsman, Ilharco,
  Gordon, Schuhmann, Schmidt, and Jitsev}]{cherti2023reproducible}
Cherti, M.; Beaumont, R.; Wightman, R.; Wortsman, M.; Ilharco, G.; Gordon, C.;
  Schuhmann, C.; Schmidt, L.; and Jitsev, J. 2023.
\newblock Reproducible Scaling Laws for Contrastive Language-Image Learning.
\newblock In \emph{Proceedings of the IEEE/CVF Conference on Computer Vision
  and Pattern Recognition}, 2818--2829.

\bibitem[{Deng et~al.(2009)Deng, Dong, Socher, Li, Li, and
  Fei-Fei}]{deng2009imagenet}
Deng, J.; Dong, W.; Socher, R.; Li, L.-J.; Li, K.; and Fei-Fei, L. 2009.
\newblock ImageNet: A Large-Scale Hierarchical Image Database.
\newblock In \emph{Proceedings of the IEEE Conference on Computer Vision and
  Pattern Recognition}, 248--255.

\bibitem[{Gong et~al.(2026)Gong, Yu, Al-Nuaimy, and Xiao}]{gong2026learning}
Gong, Y.; Yu, S.; Al-Nuaimy, W.; and Xiao, J. 2026.
\newblock Learning from Itself: Mining Internal Knowledge from Vision Language
  Models for Continual Learning.
\newblock In \emph{Proceedings of the IEEE/CVF Conference on Computer Vision
  and Pattern Recognition}, 10830--10839.

\bibitem[{He et~al.(2026{\natexlab{a}})He, Qiu, Meng, Xu, Wu, and
  Li}]{he2026desclip}
He, C.; Qiu, Z.; Meng, F.; Xu, L.; Wu, Q.; and Li, H. 2026{\natexlab{a}}.
\newblock DesCLIP: Robust Continual Learning via General Attribute Descriptions
  for VLM-Based Visual Recognition.
\newblock \emph{IEEE Transactions on Multimedia}, 28: 5021--5035.

\bibitem[{He, Duan, and Zhu(2025)}]{he2025cllora}
He, J.; Duan, Z.; and Zhu, F. 2025.
\newblock CL-LoRA: Continual Low-Rank Adaptation for Rehearsal-Free
  Class-Incremental Learning.
\newblock In \emph{Proceedings of the IEEE/CVF Conference on Computer Vision
  and Pattern Recognition}, 30534--30544.

\bibitem[{He et~al.(2026{\natexlab{b}})He, Cheng, Wang, Yang, Wang, and
  Gao}]{he2026loda}
He, L.; Cheng, D.; Wang, H.; Yang, X.; Wang, N.; and Gao, X.
  2026{\natexlab{b}}.
\newblock Task-Driven Subspace Decomposition for Knowledge Sharing and
  Isolation in LoRA-Based Continual Learning.
\newblock In \emph{Proceedings of the 43rd International Conference on Machine
  Learning}, volume 306 of \emph{Proceedings of Machine Learning Research}.

\bibitem[{Hendrycks et~al.(2021)Hendrycks, Basart, Mu, Kadavath, Wang, Dorundo,
  Desai, Zhu, Parajuli, Guo, Song, Steinhardt, and Gilmer}]{hendrycks2021many}
Hendrycks, D.; Basart, S.; Mu, N.; Kadavath, S.; Wang, F.; Dorundo, E.; Desai,
  R.; Zhu, T.; Parajuli, S.; Guo, M.; Song, D.; Steinhardt, J.; and Gilmer, J.
  2021.
\newblock The Many Faces of Robustness: A Critical Analysis of
  Out-of-Distribution Generalization.
\newblock In \emph{Proceedings of the IEEE/CVF International Conference on
  Computer Vision}, 8340--8349.

\bibitem[{Horn and Johnson(2012)}]{horn2012matrix}
Horn, R.~A.; and Johnson, C.~R. 2012.
\newblock \emph{Matrix Analysis}.
\newblock Cambridge University Press, second edition.

\bibitem[{Hu et~al.(2022)Hu, Shen, Wallis, Allen-Zhu, Li, Wang, Wang, and
  Chen}]{hu2022lora}
Hu, E.~J.; Shen, Y.; Wallis, P.; Allen-Zhu, Z.; Li, Y.; Wang, S.; Wang, L.; and
  Chen, W. 2022.
\newblock LoRA: Low-Rank Adaptation of Large Language Models.
\newblock In \emph{International Conference on Learning Representations}.

\bibitem[{Huang et~al.(2024)Huang, Cao, Lu, and Liu}]{huang2024rapf}
Huang, L.; Cao, X.; Lu, H.; and Liu, X. 2024.
\newblock Class-Incremental Learning with CLIP: Adaptive Representation
  Adjustment and Parameter Fusion.
\newblock In \emph{European Conference on Computer Vision}, 214--231.

\bibitem[{Huang et~al.(2025)Huang, Cao, Lu, Meng, Yang, and
  Liu}]{huang2025mind}
Huang, L.; Cao, X.; Lu, H.; Meng, Y.; Yang, F.; and Liu, X. 2025.
\newblock Mind the Gap: Preserving and Compensating for the Modality Gap in
  CLIP-Based Continual Learning.
\newblock In \emph{Proceedings of the IEEE/CVF International Conference on
  Computer Vision}, 3777--3786.

\bibitem[{Jha, Gong, and Yao(2024)}]{jha2024clap}
Jha, S.; Gong, D.; and Yao, L. 2024.
\newblock {CLAP4CLIP}: Continual Learning with Probabilistic Finetuning for
  Vision-Language Models.
\newblock In \emph{Advances in Neural Information Processing Systems},
  volume~37, 129146--129186.

\bibitem[{Krause et~al.(2013)Krause, Stark, Deng, and Fei-Fei}]{krause2013cars}
Krause, J.; Stark, M.; Deng, J.; and Fei-Fei, L. 2013.
\newblock 3D Object Representations for Fine-Grained Categorization.
\newblock In \emph{Proceedings of the IEEE International Conference on Computer
  Vision Workshops}, 554--561.

\bibitem[{Krizhevsky(2009)}]{krizhevsky2009learning}
Krizhevsky, A. 2009.
\newblock Learning Multiple Layers of Features from Tiny Images.
\newblock Technical report, University of Toronto.

\bibitem[{Li et~al.(2026)Li, Hu, Zhou, Yang, Ye, and Zhan}]{li2025bofa}
Li, L.; Hu, T.; Zhou, D.-W.; Yang, J.-Q.; Ye, H.-J.; and Zhan, D.-C. 2026.
\newblock BOFA: Bridge-Layer Orthogonal Low-Rank Fusion for CLIP-Based
  Class-Incremental Learning.
\newblock In \emph{Proceedings of the AAAI Conference on Artificial
  Intelligence}, 22967--22975.

\bibitem[{Liang et~al.(2022)Liang, Zhang, Kwon, Yeung, and Zou}]{liang2022mind}
Liang, V.~W.; Zhang, Y.; Kwon, Y.; Yeung, S.; and Zou, J.~Y. 2022.
\newblock Mind the Gap: Understanding the Modality Gap in Multi-modal
  Contrastive Representation Learning.
\newblock In \emph{Advances in Neural Information Processing Systems},
  volume~35, 17612--17625.

\bibitem[{Liang and Li(2024)}]{liang2024inflora}
Liang, Y.-S.; and Li, W.-J. 2024.
\newblock InfLoRA: Interference-Free Low-Rank Adaptation for Continual
  Learning.
\newblock In \emph{Proceedings of the IEEE/CVF Conference on Computer Vision
  and Pattern Recognition}, 23638--23647.

\bibitem[{Marczak et~al.(2024)Marczak, Twardowski, Trzci{\'n}ski, and
  Cygert}]{marczak2024magmax}
Marczak, D.; Twardowski, B.; Trzci{\'n}ski, T.; and Cygert, S. 2024.
\newblock {MAGMAX}: Leveraging Model Merging for Seamless Continual Learning.
\newblock In \emph{European Conference on Computer Vision}, 379--395.

\bibitem[{Radford et~al.(2021)Radford, Kim, Hallacy, Ramesh, Goh, Agarwal,
  Sastry, Askell, Mishkin, Clark, Krueger, and Sutskever}]{radford2021learning}
Radford, A.; Kim, J.~W.; Hallacy, C.; Ramesh, A.; Goh, G.; Agarwal, S.; Sastry,
  G.; Askell, A.; Mishkin, P.; Clark, J.; Krueger, G.; and Sutskever, I. 2021.
\newblock Learning Transferable Visual Models from Natural Language
  Supervision.
\newblock In \emph{Proceedings of the International Conference on Machine
  Learning}, 8748--8763.

\bibitem[{Schuhmann et~al.(2021)Schuhmann, Vencu, Beaumont, Kaczmarczyk,
  Mullis, Katta, Coombes, Jitsev, and Komatsuzaki}]{schuhmann2021laion}
Schuhmann, C.; Vencu, R.; Beaumont, R.; Kaczmarczyk, R.; Mullis, C.; Katta, A.;
  Coombes, T.; Jitsev, J.; and Komatsuzaki, A. 2021.
\newblock LAION-400M: Open Dataset of CLIP-Filtered 400 Million Image-Text
  Pairs.
\newblock arXiv:2111.02114.

\bibitem[{Sharma et~al.(2018)Sharma, Ding, Goodman, and
  Soricut}]{sharma2018conceptual}
Sharma, P.; Ding, N.; Goodman, S.; and Soricut, R. 2018.
\newblock Conceptual Captions: A Cleaned, Hypernymed, Image Alt-text Dataset
  for Automatic Image Captioning.
\newblock In \emph{Proceedings of the 56th Annual Meeting of the Association
  for Computational Linguistics}, 2556--2565.

\bibitem[{Smith et~al.(2023)Smith, Karlinsky, Gutta, Cascante-Bonilla, Kim,
  Arbelle, Panda, Feris, and Kira}]{smith2023coda}
Smith, J.~S.; Karlinsky, L.; Gutta, V.; Cascante-Bonilla, P.; Kim, D.; Arbelle,
  A.; Panda, R.; Feris, R.; and Kira, Z. 2023.
\newblock CODA-Prompt: COntinual Decomposed Attention-Based Prompting for
  Rehearsal-Free Continual Learning.
\newblock In \emph{Proceedings of the IEEE/CVF Conference on Computer Vision
  and Pattern Recognition}, 11909--11919.

\bibitem[{Thengane et~al.(2022)Thengane, Khan, Hayat, and
  Khan}]{thengane2022clip}
Thengane, V.; Khan, S.; Hayat, M.; and Khan, F.~S. 2022.
\newblock CLIP Model is an Efficient Continual Learner.
\newblock arXiv:2210.03114.

\bibitem[{Wah et~al.(2011)Wah, Branson, Welinder, Perona, and
  Belongie}]{wah2011caltech}
Wah, C.; Branson, S.; Welinder, P.; Perona, P.; and Belongie, S. 2011.
\newblock The Caltech-UCSD Birds-200-2011 Dataset.
\newblock Technical Report CNS-TR-2011-001, California Institute of Technology.

\bibitem[{Wang et~al.(2023)Wang, Chen, Ge, Xia, Bao, Zheng, Zhang, Gui, and
  Huang}]{wang2023olora}
Wang, X.; Chen, T.; Ge, Q.; Xia, H.; Bao, R.; Zheng, R.; Zhang, Q.; Gui, T.;
  and Huang, X. 2023.
\newblock Orthogonal Subspace Learning for Language Model Continual Learning.
\newblock In \emph{Findings of the Association for Computational Linguistics:
  EMNLP 2023}, 10658--10671.

\bibitem[{Wang et~al.(2022{\natexlab{a}})Wang, Zhang, Ebrahimi, Sun, Zhang,
  Lee, Ren, Su, Perot, Dy, and Pfister}]{wang2022dualprompt}
Wang, Z.; Zhang, Z.; Ebrahimi, S.; Sun, R.; Zhang, H.; Lee, C.-Y.; Ren, X.; Su,
  G.; Perot, V.; Dy, J.; and Pfister, T. 2022{\natexlab{a}}.
\newblock DualPrompt: Complementary Prompting for Rehearsal-Free Continual
  Learning.
\newblock In \emph{European Conference on Computer Vision}, 631--648.

\bibitem[{Wang et~al.(2022{\natexlab{b}})Wang, Zhang, Lee, Zhang, Sun, Ren, Su,
  Perot, Dy, and Pfister}]{wang2022l2p}
Wang, Z.; Zhang, Z.; Lee, C.-Y.; Zhang, H.; Sun, R.; Ren, X.; Su, G.; Perot,
  V.; Dy, J.; and Pfister, T. 2022{\natexlab{b}}.
\newblock Learning to Prompt for Continual Learning.
\newblock In \emph{Proceedings of the IEEE/CVF Conference on Computer Vision
  and Pattern Recognition}, 139--149.

\bibitem[{Xiao et~al.(2010)Xiao, Hays, Ehinger, Oliva, and
  Torralba}]{xiao2010sun}
Xiao, J.; Hays, J.; Ehinger, K.~A.; Oliva, A.; and Torralba, A. 2010.
\newblock SUN Database: Large-Scale Scene Recognition from Abbey to Zoo.
\newblock In \emph{Proceedings of the IEEE Conference on Computer Vision and
  Pattern Recognition}, 3485--3492.

\bibitem[{Yu et~al.(2024{\natexlab{a}})Yu, Zhuge, Zhang, Hu, Wang, Lu, and
  He}]{yu2024moe4cl}
Yu, J.; Zhuge, Y.; Zhang, L.; Hu, P.; Wang, D.; Lu, H.; and He, Y.
  2024{\natexlab{a}}.
\newblock Boosting Continual Learning of Vision-Language Models via
  Mixture-of-Experts Adapters.
\newblock In \emph{Proceedings of the IEEE/CVF Conference on Computer Vision
  and Pattern Recognition}, 23219--23230.

\bibitem[{Yu et~al.(2024{\natexlab{b}})Yu, Tao, Goswami, Yao, Twardowski,
  Van~de Weijer, and Xu}]{yu2024exploiting}
Yu, L.; Tao, Z.; Goswami, D.; Yao, H.; Twardowski, B.; Van~de Weijer, J.; and
  Xu, C. 2024{\natexlab{b}}.
\newblock Exploiting the Semantic Knowledge of Pre-trained Text-Encoders for
  Continual Learning.
\newblock \emph{arXiv preprint arXiv:2408.01076}.

\bibitem[{Yu et~al.(2025)Yu, Ko, Liu, Dong, Wu, and Zhu}]{yu2025language}
Yu, Y.; Ko, S.; Liu, H.; Dong, Y.; Wu, X.; and Zhu, Q. 2025.
\newblock Language Guided Concept Bottleneck Models for Interpretable Continual
  Learning.
\newblock In \emph{Proceedings of the IEEE/CVF Conference on Computer Vision
  and Pattern Recognition}, 14976--14986.

\bibitem[{Zhang et~al.(2023)Zhang, Wang, Kang, Chen, and Wei}]{zhang2023slca}
Zhang, G.; Wang, L.; Kang, G.; Chen, L.; and Wei, Y. 2023.
\newblock {SLCA}: Slow Learner with Classifier Alignment for Continual Learning
  on a Pre-Trained Model.
\newblock In \emph{Proceedings of the IEEE/CVF International Conference on
  Computer Vision}, 19148--19158.

\bibitem[{Zhang et~al.(2024)Zhang, Janson, Aljundi, and
  Elhoseiny}]{zhang2024overcoming}
Zhang, W.; Janson, P.; Aljundi, R.; and Elhoseiny, M. 2024.
\newblock Overcoming Generic Knowledge Loss with Selective Parameter Update.
\newblock In \emph{Proceedings of the IEEE/CVF Conference on Computer Vision
  and Pattern Recognition (CVPR)}, 24046--24056.

\bibitem[{Zhao et~al.(2024)Zhao, Wang, Zeng, Zhao, and
  Chan}]{zhao2024gradeclip}
Zhao, C.; Wang, K.; Zeng, X.; Zhao, R.; and Chan, A.~B. 2024.
\newblock Gradient-Based Visual Explanation for Transformer-Based {CLIP}.
\newblock In \emph{Proceedings of the 41st International Conference on Machine
  Learning}, volume 235 of \emph{Proceedings of Machine Learning Research},
  61072--61091.

\bibitem[{Zheng et~al.(2023)Zheng, Ma, Wang, Qin, Yue, and
  You}]{zheng2023preventing}
Zheng, Z.; Ma, M.; Wang, K.; Qin, Z.; Yue, X.; and You, Y. 2023.
\newblock Preventing Zero-Shot Transfer Degradation in Continual Learning of
  Vision-Language Models.
\newblock In \emph{Proceedings of the IEEE/CVF International Conference on
  Computer Vision (ICCV)}, 19125--19136.

\bibitem[{Zhou et~al.(2025{\natexlab{a}})Zhou, Cai, Ye, Zhan, and
  Liu}]{zhou2025aper}
Zhou, D.-W.; Cai, Z.-W.; Ye, H.-J.; Zhan, D.-C.; and Liu, Z.
  2025{\natexlab{a}}.
\newblock Revisiting Class-Incremental Learning with Pre-Trained Models:
  Generalizability and Adaptivity Are All You Need.
\newblock \emph{International Journal of Computer Vision}, 133(3): 1012--1032.

\bibitem[{Zhou et~al.(2025{\natexlab{b}})Zhou, Li, Ning, Ye, Zhang, and
  Zhan}]{zhou2025engine}
Zhou, D.-W.; Li, K.-W.; Ning, J.; Ye, H.-J.; Zhang, L.; and Zhan, D.-C.
  2025{\natexlab{b}}.
\newblock External Knowledge Injection for CLIP-Based Class-Incremental
  Learning.
\newblock In \emph{Proceedings of the IEEE/CVF International Conference on
  Computer Vision}, 3314--3325.

\bibitem[{Zhou et~al.(2025{\natexlab{c}})Zhou, Zhang, Wang, Ning, Ye, Zhan, and
  Liu}]{zhou2025proof}
Zhou, D.-W.; Zhang, Y.; Wang, Y.; Ning, J.; Ye, H.-J.; Zhan, D.-C.; and Liu, Z.
  2025{\natexlab{c}}.
\newblock Learning Without Forgetting for Vision-Language Models.
\newblock \emph{IEEE Transactions on Pattern Analysis and Machine
  Intelligence}, 47(6): 4489--4504.

\end{thebibliography}
\fi
\endgroup

\ifdefined\DuLBEAppendixEmbedded
\else
\end{document}
\fi

\end{bibunit}

\end{document}